\documentclass{article}

\PassOptionsToPackage{numbers, compress}{natbib}

 \usepackage[main, final]{neurips_2026}

\usepackage[utf8]{inputenc} 
\usepackage[T1]{fontenc}    
\usepackage{hyperref}       
\usepackage{url}            
\usepackage{booktabs}       
\usepackage{amsfonts}       
\usepackage{nicefrac}       
\usepackage{microtype}      
\usepackage{xcolor}         
\usepackage{graphicx}
\usepackage{amsmath}
\usepackage{multirow}
\usepackage{enumitem}
\usepackage{subcaption}
\usepackage[capitalize,noabbrev]{cleveref}

\newcommand\blfootnote[1]{%
  \begingroup
  \renewcommand\thefootnote{}\footnote{#1}%
  \addtocounter{footnote}{-1}%
  \endgroup
}

\title{Efficient Hybrid Distillation: Synergizing Score and Adversarial Objectives for One-Step Diffusion}

\author{
Fei Peng$^{1,2}$ \quad 
Junqiang Wu$^3$ \quad 
Haoxian Tan$^3$ \quad
Jun Zhou$^1$ \quad \\
\textbf{Jie Hu}$^3$ \quad
\textbf{Xiaoming Wei}$^3$ \quad
\textbf{Jie Guo}$^2$\footnotemark[2] \quad 
\textbf{Xiu Li}$^1$\footnotemark[2]\\
$^1$Tsinghua University\quad
$^2$Peng Cheng Laboratory\quad
$^3$Meituan Inc.
}

\begin{document}
\maketitle
\blfootnote{$^\dagger$Corresponding author. Work done at Meituan.}

\vspace{-1.2cm}
\begin{figure}[h]
    \centering
    \includegraphics[width=1.0\linewidth]{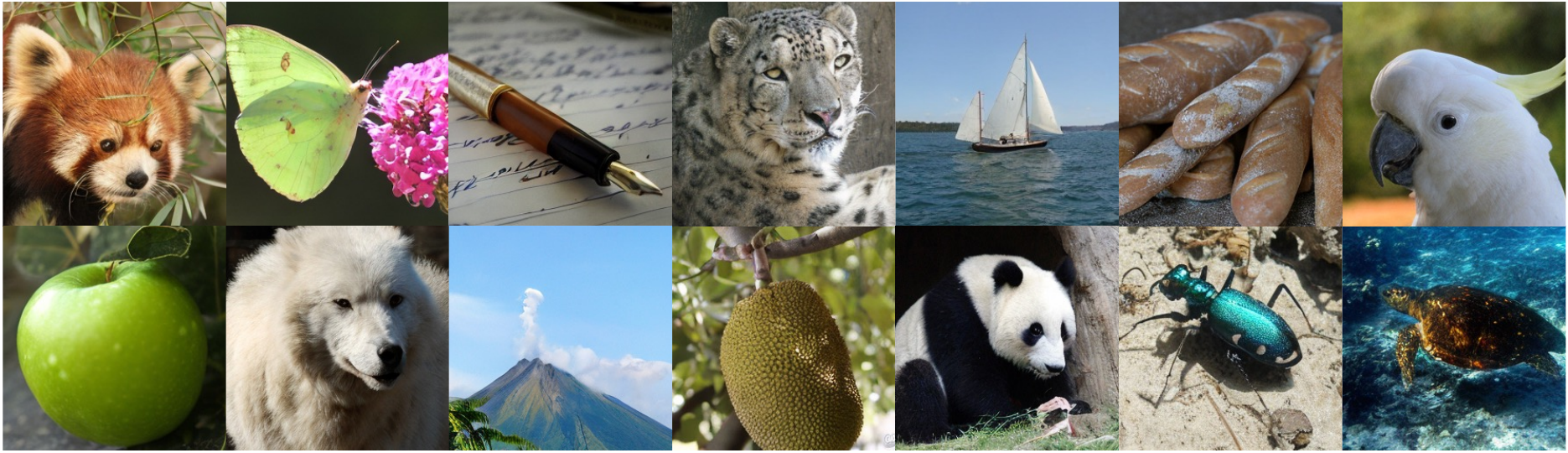}
    \caption{\textbf{Qualitative Results on ImageNet 256$\times$256.} Samples generated by our EASD-XL/2 with \textbf{one inference step}. More results are provided in \textbf{Appendix~\cref{app:more_samples}}.}
    \label{fig:qualitative_results}
\end{figure}

\begin{abstract}
Numerous distillation methods have been developed to accelerate diffusion models. Recent research indicates that hybrid strategies—blending different paradigms—yield superior results compared to single-method approaches. However, current works predominantly rely on a simple combination of objectives, leaving the question of how to effectively synergize them largely underexplored. In this paper, we specifically investigate the synergy between Score Distillation and Adversarial Distillation. We reveal that their operating zones are distinct and complementary: Score Distillation is effective in high-noise regimes for capturing the global distribution yet lacks fine-grained supervision in low-noise regions. Conversely, Adversarial Distillation is effective in low-noise regions but is prone to instability and suffers from discriminator failure when applied to high-noise intervals. Leveraging this insight, we propose Efficient Adversarial Score Distillation (EASD), a framework that synergizes these objectives via a regime-adaptive sampling strategy, directing each objective to focus on its effective interval. We further apply this paradigm to DiT-based Flow Matching models. Empirically, our method achieves a promising one-step FID of 1.39 on ImageNet 256$\times$256 using the SiT-XL/2+REPA model. This work provides a concrete guideline for maximizing the potential of hybrid distillation. Our code and model are available at \href{https://github.com/pf0607/EASD}{https://github.com/pf0607/EASD}.
\end{abstract}
\section{Introduction}

Diffusion models \cite{ho2020denoising, song2020score} have revolutionized generative AI, enabling the synthesis of high-fidelity images \cite{esser2024scaling,labs2025flux1kontextflowmatching,rombach2022high,podell2023sdxl,wu2025qwen,cai2025z,xu2025hunyuanportrait,zhou2025fireedit} and videos \cite{ yang2024cogvideox, wan2025wan,gupta2024photorealistic,bar2024lumiere} with unprecedented quality. However, their reliance on iterative sampling processes—typically requiring tens to hundreds of sequential function evaluations—imposes significant latency and computational burdens, hindering real-time applications. Consequently, accelerating diffusion models has become a paramount research focus.

To address this efficiency bottleneck, various distillation paradigms have emerged. \textbf{Consistency Models} \cite{song2023consistency, luo2023latent,kim2023consistency,lu2024simplifying} learn to map noise directly to data by enforcing self-consistency. \textbf{Adversarial Distillation} methods \cite{xu2024ufogen,sauer2024fast} leverage GAN \cite{goodfellow2014generative} objectives with pre-trained teacher discriminators to ensure perceptual fidelity in few-step generation. \textbf{Score Distillation} approaches \cite{yin2024one,zhou2024score} minimize distribution divergence via score matching gradients, preserving diversity and geometric structure.

Recent studies indicate that \textbf{hybrid strategies} combining these distinct paradigms can unlock superior performance. For instance, DMD2 \cite{yin2024improved} and SiDA \cite{zhou2024adversarial} integrate Adversarial Distillation with Score Distillation. 
Other works like FreeFlow \cite{tong2025flow} and rCM \cite{zheng2025large} also explore combining Consistency Model with Score Distillation. Despite the empirical success of these combinations, the theoretical understanding of \textit{why} and \textit{how} they complement each other remains under-explored. Specifically, existing works often lack a granular analysis of the temporal effectiveness of each objective, leading to suboptimal combinations that may suffer from training instability or mode collapse.

In this paper, we present \textbf{Efficient Adversarial Score Distillation (EASD)}, a framework designed to maximize the synergy between adversarial and score objectives. We structure our contributions as follows:
1) We provide a theoretical unification and empirical decoupling of Score Distillation and Adversarial Distillation. While we establish that both methods share the same underlying optimization goal—indirect distribution matching—which explains why they can be combined, our analysis further reveals their opposing yet complementary effective and ineffective timestep regimes: Score Distillation operates efficiently at high-noise timesteps but degrades at low-noise timesteps, while Adversarial Distillation exhibits the opposite behavior, which explains why their combination yields improved performance.
2) We propose Regime-Adaptive Sampling, which encourages each distillation objective to operate within its complementary effective timestep region, improving the overall efficiency of hybrid distillation. Building on this, we further extend the hybrid Adversarial and Score Distillation strategy to modern scalable architectures, formulating the Score Distillation objective within the Flow Matching framework and designing a streamlined Frozen-Teacher Discriminator tailored for Diffusion Transformers (DiT). By integrating these components, we establish a simple and effective baseline for combining these distillation methods on state-of-the-art generative backbones.
3) We introduce comprehensive mechanisms to stabilize training and mitigate mode collapse. We propose techniques such as Fake Score EMA and Iterative Warmup to ensure robust convergence. Furthermore, we analyze the ``compounding effect'' of Classifier-Free Guidance in distillation and introduce a Latent Mode Seeking loss. This allows us to navigate the trade-off between fidelity and diversity.

Empirically, EASD achieves a state-of-the-art FID of \textbf{1.39} on ImageNet $256 \times 256$ using the SiT-XL/2+REPA \cite{yu2024representation} backbone with only 60 training epochs, surpassing recent baselines and providing a concrete guideline
for maximizing the potential of hybrid distillation..
\section{Related Work}

\noindent \textbf{Training-Free Acceleration.}
Early efforts focused on improving the discretization of the reverse ODE/SDE without retraining. Besides DDIM \cite{song2020denoising}, high-order ODE solvers such as PNDM \cite{liu2022pseudo}, DPM-Solver++ \cite{lu2022dpm,lu2025dpm}, and DEIS \cite{zhang2022fast} have pushed the limit of training-free sampling to around 10--20 steps. However, quality degrades significantly at extreme compression (1--4 steps).

\noindent \textbf{Consistency and Flow Trajectory Mapping.}
\textit{Consistency Models} \cite{song2023consistency} enforce self-consistency along the probability trajectory to enable few-step generation. This paradigm has been extended to latent spaces via LCM \cite{luo2023latent} and refined by trajectory-based variants such as CTM \cite{kim2023consistency}, tCM \cite{song2023improved}, sCM \cite{lu2024simplifying}, and PCM \cite{wang2024phased}.
Parallel to this, \textit{Flow Matching} approaches, including Rectified Flow \cite{liu2022flow}, InstaFlow \cite{liu2023instaflow}, and PeRFlow \cite{yan2024perflow}, aim to straighten generative ODE trajectories (Reflow) to minimize discretization error. While effective, these methods often require complex training objectives (e.g., self-differentiation) or specialized solver-based distillation.

\noindent \textbf{Adversarial Distillation.}
To achieve few-step generation with high perceptual fidelity, Adversarial Distillation \cite{lin2024sdxl,sauer2024adversarial} integrates GAN \cite{goodfellow2014generative} objectives into diffusion training. Methods such as UFOGen \cite{xu2024ufogen} and LADD \cite{sauer2024fast} utilize the pre-trained diffusion backbone as a discriminator. By discriminating on renoised latents or features, these methods force the student to match the teacher's perceptual distribution. While effective at generating sharp textures, adversarial approaches are inherently prone to training instability and mode collapse, often sacrificing diversity for visual quality.

\noindent \textbf{Score Distillation.}
Score Distillation \cite{luo2023diff,shaovarflow,luo2024one} aims to minimize the divergence between student and teacher distributions without explicit adversarial gaming. Rooted in Score Distillation Sampling (SDS) \cite{poole2022dreamfusion} and Variational Score Distillation (VSD) \cite{wang2023prolificdreamer}, methods like DMD \cite{yin2024one} and SiD \cite{zhou2024score} employ the score difference between the teacher and a dynamically trained student as a gradient to update the generator. Recent work like SMD \cite{jayashankar2025score} further explores mixture distributions for better alignment. These approaches effectively preserve the geometric structure of the data manifold and mitigate mode collapse but may lack the absolute sharpness of GAN-based methods in low-noise regions.

\noindent \textbf{Hybrid Strategies.}
Recognizing the complementary strengths of different paradigms, recent research has shifted towards hybrid distillation frameworks. For instance, DMD2 \cite{yin2024improved}, SiDA \cite{zhou2024adversarial}, f-distill \cite{xu2025one} and others \cite{wei2024adversarial, lu2025adversarial} integrate the score distillation with adversarial distillation. Similarly, rCM \cite{zheng2025large}, FreeFlow \cite{tong2025flow} and TurboDiffusion \cite{zhang2025turbodiffusion}. Despite the empirical success of these combinations, the theoretical understanding of their synergy—specifically how different loss components and sampling strategies interact to balance fidelity and diversity—remains under-explored. 
\section{Preliminaries}

\subsection{Diffusion Models}
\label{sec:prelim_diffusion}
Diffusion models generate data by reversing a forward process that maps data $x_0 \sim p_{\text{data}}$ to noise $x_1 \sim \mathcal{N}(0, I)$. 
While standard diffusion models often operate on discrete timesteps, Flow Matching is defined over a continuous time horizon $t \in [0, 1]$.
The forward transition $q(x_t|x_0)$ is defined as:
\begin{equation}
    x_t = \alpha_t x_0 + \sigma_t \epsilon, \quad \epsilon \sim \mathcal{N}(0, I).
    \label{eq:forward}
\end{equation}
This formulation unifies various schedules, including VP-SDE \cite{song2020score} and Rectified Flow \cite{liu2022flow} (where $\alpha_t=1-t, \sigma_t=t$).
To learn the data distribution, a time-dependent network $F_\phi(x_t, t)$ is trained to regress a target signal $y_t$ from the noisy state $x_t$. The universal training objective is:
\begin{equation}
    \mathcal{L}_{\text{diff}}(\phi) = \mathbb{E}_{t, x_0, \epsilon} \left[ \| F_\phi(x_t, t) - y_t \|^2 \right].
    \label{eq:diff_loss}
\end{equation}
The target $y_t$ varies by parameterization: 
$y_t=\epsilon$ corresponds to the noise prediction objective used in DDPM \cite{ho2020denoising} and Score-based SDEs \cite{song2020score};
$y_t=v_t=\epsilon-x_0$ corresponds to \textbf{Flow Matching} (specifically Rectified Flow \cite{liu2022flow}); 
and $y_t=x_0$ corresponds to \textbf{EDM} \cite{karras2022elucidating}.

\subsection{Adversarial Distillation}
Adversarial distillation \cite{sauer2024adversarial,sauer2024fast} distills generator $G_\theta$ against a discriminator $D_\psi$. 
$D_\psi$ distinguishes between real samples $x_0$ and generated samples based on their noisy latents perturbed via \cref{eq:forward}:
\begin{equation}
    \hat{x}_t = \alpha_t G_\theta(z) + \sigma_t \epsilon, \quad z \sim \mathcal{N}(0, I).
    \label{eq:forward_generated}
\end{equation}

\noindent \textbf{Discriminator Update:} $D_\psi$ minimizes the classification error over the joint distribution:
\begin{equation}
    \mathcal{L}_D(\psi) = \mathbb{E}_{x_0, t, \epsilon} \left[ -\log D_\psi(x_t) \right] +\mathbb{E}_{z, t, \epsilon} \left[ -\log(1 - D_\psi(\hat{x}_t)) \right].
\end{equation}
\noindent \textbf{Generator Update:} The student minimizes the adversarial loss to fool the discriminator:
\begin{equation}
    \mathcal{L}_{G}^{\text{adv}}(\theta) = \mathbb{E}_{z, t, \epsilon} [ -\log D_\psi(\hat{x}_t) ].
\end{equation}
\subsection{Score Distillation}
Score distillation aligns the student distribution $p_\theta$ (induced by the few-step generator $G_\theta$) with the teacher distribution $p_{\text{real}}$ by minimizing the Reverse KL divergence.
The gradient corresponds to the difference between their score functions:
\begin{equation}
    \nabla_x D_{\text{KL}}(p_\theta \| p_{\text{real}}) = \nabla_x \log p_\theta(x) - \nabla_x \log p_{\text{real}}(x).
\end{equation}
Direct computation of these scores is intractable. \cite{song2020score} demonstrated that optimizing the denoising objective (\cref{eq:diff_loss}) is equivalent to score matching, implying that the model implicitly learns the score function (specifically, $\nabla \log p_t(x_t) \approx -\epsilon / \sigma_t$).
Leveraging this relationship, we approximate the \textbf{Real Score} $\nabla_x \log p_{\text{real}}$ using the pre-trained teacher diffusion model (denoted as $s_{\text{real}}$). 
Simultaneously, to estimate the \textbf{Student Score} $\nabla_x \log p_\theta$, we train a separate diffusion network $F_\phi$ on generated samples using \cref{eq:diff_loss} (denoted as $s_{\text{fake}}$).
The generator parameter update is derived via a variational decomposition \cite{wang2023prolificdreamer}:
\begin{equation}
    \nabla_\theta \mathcal{L}_{\text{SD}} \approx \underset{z, t, \epsilon}{\mathbb{E}} \left[ w_t (s_{\text{fake}}(\hat{x}_t,t) - s_{\text{real}}(\hat{x}_t,t)) \nabla_\theta G_\theta(z) \right].
    \label{eq:dmd_grad}
\end{equation}
$s_{\text{fake}}$ and $s_{\text{real}}$ are derived from the outputs of $F_\phi$ and the teacher model. $\hat{x}_t$ is same in \cref{eq:forward_generated}.
\begin{figure*}[t]
    \centering
    \begin{minipage}[t]{0.49\textwidth}
        \centering
        \includegraphics[width=\textwidth]{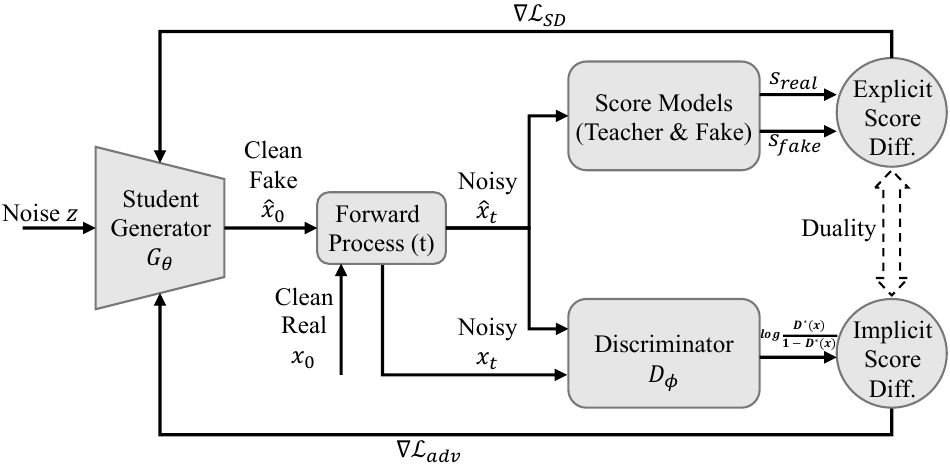}
        \caption{\textbf{Unified View of Distillation Objectives.}
        Score Distillation \textit{explicitly} regresses $(s_{\text{real}} - s_{\text{fake}})$ via a ``Composite Discriminator'';
        Adversarial Distillation \textit{implicitly} estimates it via discriminator logit gradients.
        Both optimize the same underlying objective via different proxies.}
        \label{fig:duality_diagram}
    \end{minipage}
    \hfill
    \begin{minipage}[t]{0.50\textwidth}
        \centering
        \vspace{-90pt}
        \captionof{table}{\textbf{Ablation of Sampling Strategies.} Comparison of different sampling ranges for Score Distillation (\textbf{SD}) and Adversarial Distillation (\textbf{AD}). $\mathcal{U}$ denotes Uniform sampling. Each distillation is conducted independently without combination.}
        \label{tab:ablation_sampling_combined}
        \begin{minipage}[t]{0.49\textwidth}
            \centering
            \subcaption{\textbf{SD}}
            \label{tab:score_ablation}
            \resizebox{\textwidth}{!}{
                \begin{tabular}{cc}
                    \toprule
                    Sampling Range & FID $\downarrow$ \\
                    \midrule
                    $\mathcal{U}(0.0, 1.0)$ & 2.29 \\
                    $\mathcal{U}(0.1, 1.0)$ & 2.22 \\
                    $\mathcal{U}(0.2, 1.0)$ & \textbf{2.18} \\
                    $\mathcal{U}(0.3, 1.0)$ & 2.45 \\
                    $\mathcal{U}(0.4, 1.0)$ & 3.18 \\
                    $\mathcal{U}(0.0, 0.9)$ & 2.32 \\
                    $\mathcal{U}(0.0, 0.8)$ & 2.60 \\
                    \bottomrule
                \end{tabular}
            }
        \end{minipage}
        \hfill
        \begin{minipage}[t]{0.49\textwidth}
            \centering
            \subcaption{\textbf{AD}}
            \label{tab:adv_ablation}
            \resizebox{\textwidth}{!}{
                \begin{tabular}{cc}
                    \toprule
                    Sampling Range & FID $\downarrow$ \\
                    \midrule
                    $\mathcal{U}(0.0, 1.0)$ & 2.43 \\
                    $\mathcal{U}(0.1, 1.0)$ & 2.96 \\
                    $\mathcal{U}(0.2, 1.0)$ & 4.65 \\
                    $\mathcal{U}(0.3, 1.0)$ & 9.98 \\
                    $\mathcal{U}(0.0, 0.9)$ & 2.39 \\
                    $\mathcal{U}(0.0, 0.8)$ & \textbf{2.31} \\
                    $\mathcal{U}(0.0, 0.7)$ & Fail \\
                    \bottomrule
                \end{tabular}
            }
        \end{minipage}
    \end{minipage}
    \vspace{-14pt}
\end{figure*}

\section{Unified and Decoupled Perspective on Score and Adversarial Distillation}
\label{sec:analysis}
\subsection{The Duality of Indirect Distribution Matching}
\label{subsec:duality}
Despite their distinct formulations, we argue that Score Distillation and Adversarial Distillation are fundamentally connected. We elucidate this duality by interpreting the interaction between the networks as specific forms of adversarial games, revealing that \textbf{both} methods aim to minimize distribution divergence by estimating the \textbf{score difference}.

\noindent \textbf{Adversarial Interpretation of Score Distillation.}
In Score Distillation, the training objective is to minimize the difference between the student's score and the teacher's. Since the student's distribution $p_\theta$ is implicit, we cannot directly compute its score.
We propose that the combination of the frozen teacher score $s_{\text{real}}$ and the learnable student score $s_{\text{fake}}$ functions as a ``\textbf{Composite Discriminator}".
\begin{itemize}[leftmargin=*, itemsep=2pt, topsep=0pt, parsep=0pt]
    \item \textbf{Discriminator Step (Criticizing):} The $s_{\text{fake}}$ network is trained via Score Matching to estimate $\nabla \log p_\theta$. Its goal is to accurately model the student's current distribution to ``expose'' the discrepancy $(s_{\text{real}} - s_{\text{fake}})$ relative to the teacher.
    \item \textbf{Generator Step (Fooling):} The student generator $G_\theta$ updates its parameters to minimize the norm of this difference. This is equivalent to ``fooling'' the composite discriminator into believing that the student's score field is indistinguishable from the teacher's.
\end{itemize}
\noindent \textbf{Score-Matching Interpretation of Adversarial Distillation.}
Conversely, the explicit discriminator $D_\psi$ in Adversarial Distillation can be viewed as an implicit estimator of the score difference.
While the original GAN objective globally minimizes the Jensen-Shannon divergence, standard implementations typically employ the Non-Saturating Loss ($-\log D(G(z))$) to avoid gradient vanishing \cite{goodfellow2014generative}.
Crucially, theoretical analysis suggests that the local update direction provided by this loss also approximates the gradient of the Reverse KL divergence \cite{arjovsky2017towards}.
Assuming an optimal discriminator $D^*(x) = \frac{p_{\text{data}}}{p_{\text{data}} + p_\theta}$, the gradient passed to the generator is proportional to the gradient of the logit (\textbf{Appendix \cref{app:gan_score_derivation}}):
\begin{equation}
        \nabla_x \mathcal{L}_{G}^{\text{adv}}(\theta) \approx \nabla_x \log \frac{D^*(x)}{1 - D^*(x)} = \nabla_x \log p_{\text{data}}(x) - \nabla_x \log p_\theta(x).
    \label{eq:gan_score_relation}
\end{equation}
Strikingly, the right-hand side of \cref{eq:gan_score_relation} is equivalent to the score difference term $(s_{\text{real}} - s_{\text{fake}})$ used in Score Distillation. 
This reveals a fundamental \textbf{theoretical duality}: Score Distillation \textit{explicitly} minimizes the score difference via regression, whereas Adversarial Distillation \textit{implicitly} eliminates this discrepancy through density ratio estimation. Both methods thus share the \textbf{same underlying optimization goal}, which explains \textbf{why they can be combined without conflict}.

\begin{figure}
    \centering
    \includegraphics[width=0.75\linewidth]{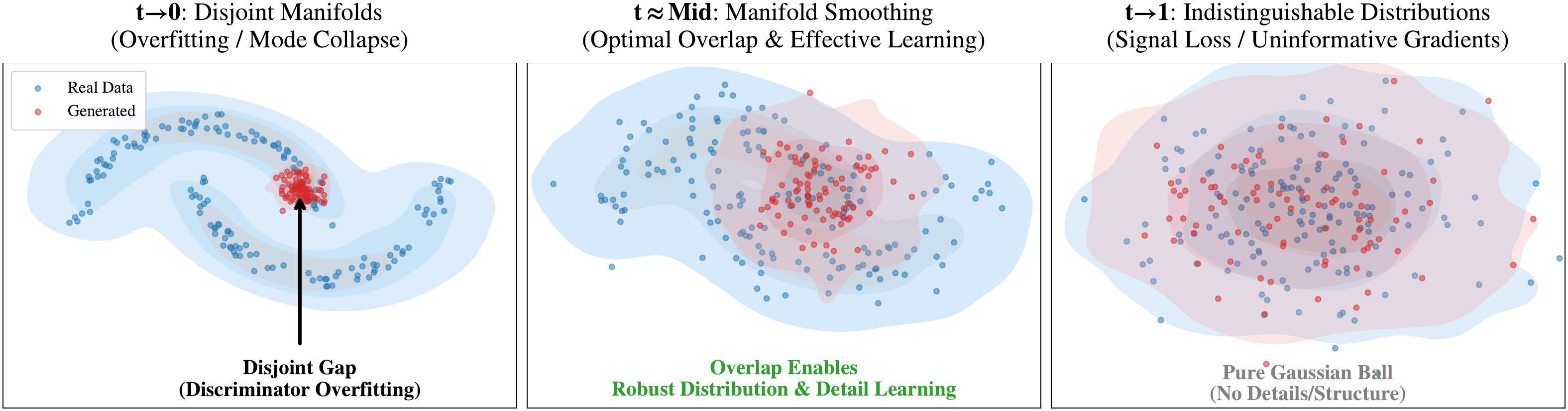}
    \caption{\textbf{Visual Analysis of Adversarial Distillation Regimes.} 
    \textbf{(Left) Low Noise ($t \to 0$):} Due to disjoint manifolds, the discriminator overfits, forcing the generator to collapse to few modes (red clump) (\textbf{Mode Collapse}).
    \textbf{(Center) Effective Regime ($t \approx \text{Mid}$):} Noise injection functions as \textbf{Manifold Smoothing}, creating optimal overlap for gradient flow while preserving \textbf{fine-grained details and roubust distribution}.
    \textbf{(Right) High Noise ($t \to 1$):} The distributions merge into an indistinguishable Gaussian ball, causing the discriminator to lose all guidance (\textbf{Signal Loss}).}
    \label{fig:manifold_smoothing}
    \vspace{-14pt}
\end{figure}

\subsection{Decoupling Effective Regimes}
\label{subsec:regimes}
Having established that Score Distillation and Adversarial Distillation share the same underlying optimization goal, a natural question arises: \textit{why does their combination yield improved performance over either method alone?} To answer this, we decouple and independently analyze the behavior of each method across different timestep regimes. We conduct ablation studies using a restricted sampling strategy $\mathcal{U}(t_{\min}, t_{\max})$, where the timestep $t$---used to perturb the student's generated image $\hat{x}_0$ via the forward process---is sampled uniformly from $[t_{\min}, t_{\max}]$. Based on \cref{tab:ablation_sampling_combined} and \cref{fig:manifold_smoothing}, we analyze the mechanism and limitations of each objective.

\textbf{Limitations of Score Distillation.}
\begin{itemize}[leftmargin=*, itemsep=2pt, topsep=0pt, parsep=2.5pt]
    \item \textbf{Low Noise Regime ($t \to 0$): Vanishing Signal vs. Necessity.} 
    As $t \to 0$, the input $\hat{x}_t$ approaches the clean data manifold. Since both the teacher and the \textbf{student's fake score model} are typically well-trained denoisers, their predicted denoising directions \textit{align closely} as the task simplifies to removing infinitesimal noise. Consequently, the gradient magnitude $\|\nabla \mathcal{L}_{\text{SD}}\|$ naturally diminishes
    , becoming susceptible to numerical instability.
    \cref{tab:score_ablation} reflects this trade-off: excluding the lowest noise interval ($\mathcal{U}(0.2, 1.0)$) yields better FID of \textbf{2.18}, outperforming the full range (FID 2.29), which suggests the extreme low-noise region ($t < 0.2$) introduces more instability than utility. However, further truncating the range ($\mathcal{U}(0.3, 1.0)$) degrades performance to FID 2.45. This indicates that the low-noise regime is indeed necessary for fine-grained detail refinement, yet standard Score Distillation \textbf{lacks effective supervision} mechanisms to leverage it efficiently due to the vanishing signal.
    \item \textbf{High Noise Regime ($t \to 1$): Diminishing Returns of Mean-Seeking.}
    At high noise levels, the score function provides global guidance but essentially points towards the center of mass of the data distribution (mean-seeking). 
    While sufficient high-noise coverage is critical for structural formation (as $\mathcal{U}(0.0, 0.8)$ drastically worsens FID to 2.60), the supervision from the \textit{extreme} high-noise regime appears to offer diminishing returns. Specifically, \cref{tab:score_ablation} shows that removing the highest noise interval ($\mathcal{U}(0.0, 0.9)$) results in only a marginal performance drop (FID 2.32 vs. 2.29 for $\mathcal{U}(0.0, 1.0)$). This suggests that while mean-seeking behavior initiates the layout, its contribution saturates as $t \to 1$, offering limited help for the specific characteristics of the generated sample.
\end{itemize}
\textbf{Limitations of Adversarial Distillation.}
\begin{itemize}[leftmargin=*, itemsep=2pt, topsep=0pt, parsep=2.5pt]
    \item \textbf{High Noise Regime ($t \to 1$): Signal Loss.}
    When $\hat{x}_t$ approaches pure noise, the perturbed real and generated distributions become statistically indistinguishable. Visually, \textbf{\cref{fig:manifold_smoothing} (Right)} shows both distributions merging into an identical Gaussian ball. In this regime, the discriminator fails to discriminate structure effectively, leading to uninformative gradients. 
    This failure is quantified in \cref{tab:adv_ablation}: restricting training to high-noise regimes (e.g., $\mathcal{U}(0.3, 1.0)$) results in a catastrophic performance drop to FID \textbf{9.98}, confirming that Adversarial Distillation is ineffective at high $t$.
    
    \item \textbf{Low Noise Regime ($t \to 0$): Overfitting and Mode Collapse.}
    As $t \to 0$, the perturbed data manifold remains extremely complex and disjoint from the generated distribution. As illustrated in \textbf{\cref{fig:manifold_smoothing} (Left)}, this \textbf{disjoint gap} allows the discriminator to overfit to trivial artifacts or memorize specific real samples. Consequently, the generator collapses onto a few specific modes to satisfy the discriminator, resulting in severe \textbf{Mode Collapse}.
    
    \item \textbf{The ``Sweet Spot": Robust Distribution Learning \& Detail Refinement.}
    Between the extremes of signal loss and overfitting, the \textit{moderately low noise regime} emerges as the optimal operating zone. In this interval (e.g., $\mathcal{U}(0.0, 0.8)$), noise injection acts as a \textbf{Manifold Smoothing} mechanism. 
    \textbf{\cref{fig:manifold_smoothing} (Center)} demonstrates this effect: the expanded supports create necessary overlap to prevent overfitting. Crucially, unlike the high-noise regime, the noise magnitude here remains sufficiently low to preserve fine-grained details. This allows the discriminator to enforce a robust boundary that facilitates both effective distribution alignment and high-frequency texture refinement, achieving better FID of \textbf{2.31}.
\end{itemize}

\noindent \textbf{Complementary Synergy.}
Our analysis reveals a clear temporal dichotomy between the two objectives: their effective and ineffective timestep regimes are precisely opposing and complementary, which explains \textbf{why their combination yields improved distillation performance over either method alone}. However, existing hybrid approaches directly combine the two objectives without considering the timestep sampling strategy, leading to inefficient resource allocation where each method wastes considerable capacity operating in its own ineffective regime. This motivates a more principled strategy that biases each distillation objective toward its complementary effective regime, maximizing synergy and improving the overall efficiency of hybrid distillation.
\vspace{-4pt}
\section{Efficient Adversarial Score Distillation}
\label{sec:method}
We propose EASD, a framework optimizing the synergy of adversarial and score objectives. It features a regime-adaptive sampling strategy for Flow Matching and DiT architectures, alongside stabilization and regularization techniques to ensure robust, diversity-preserving distillation.
\vspace{-3pt}
\subsection{Regime-Adaptive Sampling Strategy}
\label{subsec:sampling}
Building on the complementary temporal dichotomy identified in \cref{subsec:regimes}, we establish a general guideline for hybrid distillation: \textit{encouraging each distillation objective to operate more within its complementary effective regime improves the overall efficiency of hybrid distillation}. This principle can be instantiated in various ways—hard truncation of the sampling interval or soft biasing via skewed distributions both yield consistent improvements, as validated in \cref{subsec:ablation}.

In this work, we adopt a soft biasing approach using Logit-Normal distributions as a practical and transferable instantiation of this guideline. We sample the timestep $t$ by applying the sigmoid function to a normally distributed variable:
\begin{equation}
    t = \sigma(y), \quad y \sim \mathcal{N}(\mu, s^2).
    \label{eq:logit_normal_sampling}
\end{equation}
The hyperparameters $\mu$ and $s$ control the bias (center) and spread of the distribution, respectively. Guided by the regime analysis in \cref{subsec:regimes}, we configure Score Distillation with $(\mu_{\text{sd}} = 0.8, s_{\text{sd}} = 1.6)$ to bias sampling toward the high-noise regime, and Adversarial Distillation with $(\mu_{\text{adv}} = -0.8, s_{\text{adv}} = 1.6)$ to bias sampling toward the low-noise regime. The wide spread ($s=1.6$) ensures that neither objective is strictly excluded from the full timestep range, preserving the regularization benefits of residual cross-regime coverage. This configuration demonstrates strong transferability across datasets and architectures, as shown in \cref{subsec:ablation,app:transferability,app:ablation}.

\subsection{Adapting Score and Adversarial Distillation to Flow Matching and DiT}
\label{subsec:sd_for_fm}

\noindent \textbf{Score Distillation for Flow Matching.}
We adapt the score distillation objective to the Flow Matching framework.
In the original DMD implementation \cite{yin2024one}, the score functions are derived from an EDM \cite{karras2022elucidating} network $F_\phi(\cdot)$ that predicts the clean data $x_0$. 
Based on the score matching formulation \cite{song2020score}, the score function can be computed from the model's prediction $F_\phi(x_t, t)$ as:
\begin{equation}
    s(x_t,t) = - \left[x_t - \alpha_t F_\phi(x_t,t)\right]/{\sigma_t^2}.
    \label{eq:score_edm}
\end{equation}
To simplify optimization, DMD sets the weighting function $w_t = \sigma_t^2 / \alpha_t$ to counteract the coefficients. 
Substituting \cref{eq:score_edm} into the general gradient \cref{eq:dmd_grad}, the coefficients cancel out, reducing the objective to a direct regression in the $x_0$-space. The gradient of the loss is expressed as:
\begin{equation}
     \nabla_\theta \mathcal{L}_{\text{DMD}} = \mathbb{E}_{z, t} \left[ \frac{\sigma_t^2}{\alpha_t} (s_{\text{fake}} - s_{\text{real}}) \nabla_\theta G_\theta(z) \right] = \mathbb{E}_{z, t} \left[ (F_{\phi} - F_{\text{teacher}}) \nabla_\theta G_\theta(z) \right].
\label{eq:x0_diff}
\end{equation}
In the Flow Matching framework (Rectified Flow), the model predicts the velocity field $v_t$. Under the linear schedule $x_t = (1-t)x_0 + t x_1$, we have:
\begin{equation}
    x_0 = x_t - t \cdot v_t.
    \label{eq:x0_v_relation}
\end{equation}
By substituting \cref{eq:x0_v_relation} into \cref{eq:x0_diff}, we can express the gradient in terms of velocity:
\begin{equation}
    F_{\phi} - F_{\text{teacher}} = (x_t - t v_{\text{fake}}) - (x_t - t v_{\text{real}}) = -t \cdot (v_{\text{fake}} - v_{\text{real}}).
\end{equation}
Here, the coefficient $t$ emerges as a scaling factor, indicating that distilling in the $x_0$-space is equivalent to distilling in the $v$-space weighted by $t$. Since the native output of Flow Matching models is $v$, we opt to perform distillation directly in the velocity space. We discard the implicit weighting factor $t$ to stabilize training, leading to our simplified flow matching score distillation loss:
\begin{equation}
    \nabla_\theta \mathcal{L}_{\text{FSD}} = \underset{z, t, \epsilon}{\mathbb{E}} \left[ (v_{\text{fake}}(\hat{x}_t,t) - v_{\text{real}}(\hat{x}_t,t)) \nabla_\theta G_\theta(z) \right].
\end{equation}
More discussions on the choice of prediction targets are provided in \textbf{Appendix \cref{app:prediction_targets}}.

\begin{figure*}[t]
    \centering
    \begin{minipage}[t]{0.45\textwidth}
        \centering
        \includegraphics[width=\textwidth]{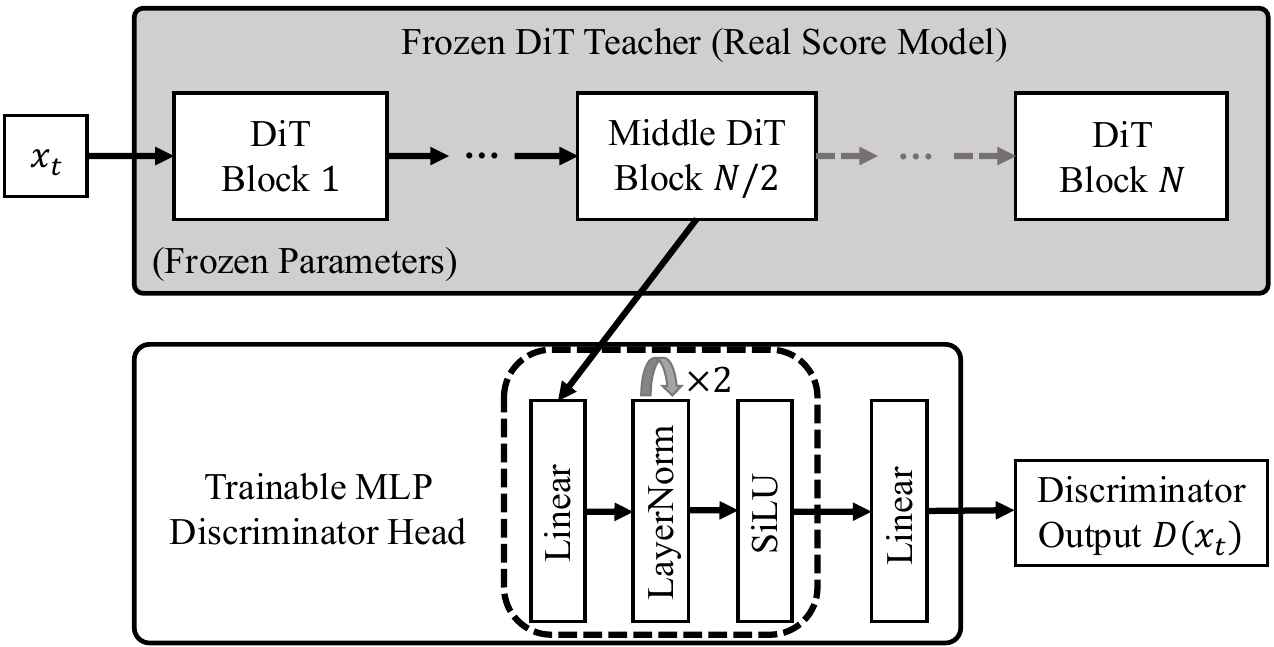}
        \caption{\textbf{Discriminator for DiT.} 
        To ensure training stability and efficiency, 
        we attach lightweight, trainable MLP heads to the middle blocks of the \textbf{frozen} DiT teacher. 
        }
        \label{fig:discriminator_arch}
    \end{minipage}
    \hfill
    \begin{minipage}[t]{0.54\textwidth}
        \centering
        \vspace{-92pt} 
        \captionof{table}{\textbf{Component Analysis.} Comparison of discriminator architectures and stabilization \& regularization techniques. ``Fake Score'' denotes attaching the discriminator head to the fake score network, and ``Block $N$'' denotes attaching it to the $N$-th DiT block of the real score network.}
        \label{tab:ablation_components}
        \begin{minipage}[t]{0.49\linewidth}
            \centering
            \subcaption{\textbf{Discriminator Design}}
            \resizebox{0.95\linewidth}{!}{
                \begin{tabular}{lc}
                    \toprule
                    Design Choice & FID $\downarrow$ \\
                    \midrule
                    Frozen Teacher (Ours) & \textbf{1.47} \\
                    Fake Score & 1.85 \\
                    Frozen Teacher Block 7 & 1.55 \\
                    Frozen Teacher Block 21 & 1.70 \\
                    2D Projection & 1.55 \\
                    \bottomrule
                \end{tabular}
            }
        \end{minipage}
        \hfill
        \begin{minipage}[t]{0.49\linewidth}
            \centering
            \subcaption{\textbf{Stabilization \& Reg.}}
            \resizebox{0.95\linewidth}{!}{
                \begin{tabular}{lc}
                    \toprule
                    Configuration~~~~~~~~~~~~~~~~~~ & FID $\downarrow$ \\
                    \midrule
                    Baseline & 1.60 \\
                    \quad + Fake Score EMA & 1.56 \\
                    \quad + Warmup 5K & 1.57 \\
                    \quad + LMS Loss & 1.58 \\
                    w/o Student EMA & 1.90 \\
                    \bottomrule
                \end{tabular}
            }
        \end{minipage}
    \end{minipage}
    \vspace{-14pt}
\end{figure*}

\noindent \textbf{Frozen-Teacher Discriminator for DiT.}
For the adversarial objective, designing an efficient discriminator for Diffusion Transformers (DiT) presents challenges compared to standard U-Net architectures. While prior methods \cite{sauer2024adversarial} often rely on heavy external Vision Foundation Models operating in pixel space, or reuse the non-stationary encoder of the updating student \cite{yin2024improved,zhou2024adversarial}, we propose a simple yet effective design tailored for DiT. We construct the discriminator by attaching lightweight, trainable MLP heads directly to the middle DiT blocks of the \textbf{frozen teacher} network (\cref{fig:discriminator_arch}). This approach offers two key advantages: 1) \textbf{Stability}: the frozen teacher provides stationary feature representations, stabilizing the adversarial game; 2) \textbf{Efficiency}: it incurs negligible computational overhead, as the teacher is already resident in memory for calculating the Score Distillation loss.

\subsection{Stabilizing the Distillation}
\label{subsec:stabilization}

To ensure the adversarial game (\cref{subsec:duality}) between the generator and the score estimator remains balanced, we introduce two techniques to stabilize the training of the fake score network $F_\phi$.

\textbf{EMA for the Fake Score Network.}
While Exponential Moving Average (EMA) is typically applied to generators, we find it also useful for the fake score network $F_\phi$. Training $F_\phi$ involves a trade-off: it must adapt quickly to the shifting student distribution (timeliness) while maintaining robust coverage. A rapidly updating $F_\phi$ may overfit to the most recent batch, providing noisy gradients.
To mitigate this, we maintain an EMA copy of $F_\phi$ with small decay rate $\beta$ and use these smoothed weights $\phi_{\text{EMA}}$ to compute the distillation target. This temporal averaging prevents the generator from chasing transient fluctuations in the score estimate, efficiently stabilizing convergence.

\textbf{Warmup.}
Unlike standard discriminators that learn from scratch, the fake score network $F_\phi$ is initialized from the pre-trained teacher. Consequently, it initially outputs scores nearly identical to $s_{\text{real}}$, resulting in zero gradients. The generator must first produce samples to ``pollute'' the distribution before $F_\phi$ can learn to detect discrepancies.
To bridge this initial lag, we employ an \textbf{Iterative Warmup} strategy: freezing the generator while training only $F_\phi$ for a few iterations. Furthermore, this can be applied periodically throughout training to ensure the ``composite discriminator'' remains synchronized with the evolving student distribution.

\subsection{Strategic Trade-off: CFG vs. Latent Mode Seeking}
\label{subsec:latent_regularization}
Classifier-Free Guidance \cite{ho2022classifier} (CFG) is standard in diffusion inference; we argue that its usage in score distillation may be detrimental. Consequently, we omit CFG in our default training setup and instead introduce a dedicated regularization term to promote diversity.

\noindent \textbf{The Compounding Trap of CFG.}
\label{subsec:cfg_compounding}
Although CFG accelerates initial convergence by sharpening images, we identify a critical failure mode: the \textbf{compounding effect} (detailed in \textbf{Appendix~\cref{app:recursive_guidance}}). Since the student inherently learns the teacher's guidance, applying CFG again during distillation creates a recursive loop (``double guidance'').
In high-diversity tasks (e.g., T2I), this focusing effect may be acceptable for semantic alignment. However, in constrained domains like ImageNet, it aggressively drives the generator to a single mode. Empirically, we observe that CFG inevitably leads to severe diversity loss and FID degradation upon prolonged training.

\noindent \textbf{Latent Mode Seeking (LMS) for Diversity.}
To counteract this collapse and restore intra-class variation, we introduce the \textbf{LMS} loss, inspired by \cite{mao2019mode}. Unlike the original pixel-space formulation, we adapt it to operate directly on the generator's latent outputs:
\begin{equation}
    \mathcal{L}_{\text{LMS}} = \| z_1 - z_2 \|/\| G_\theta(z_1) - G_\theta(z_2) \|.
\end{equation}
By maximizing the ratio of latent distance to input noise distance, LMS exerts a repulsive force that actively pushes generated samples apart, preventing them from clustering into a single mode.
\textbf{Guideline:} We advocate a context-aware strategy. For alignment-heavy tasks (e.g., T2I in \cref{tab:sdxl_coco}), CFG remains beneficial. For diversity-bottlenecked tasks (e.g., ImageNet in \cref{tab:imagenet_256}), we recommend disabling CFG and employing LMS to ensure comprehensive distribution coverage.
\section{Experiments}
\label{sec:experiments}

\subsection{Experimental Setup}
\label{subsec:experiment setup}
\textbf{Models and Datasets.} 
We evaluate the proposed Efficient Adversarial Score Distillation (EASD) framework on two distinct architectures: the U-Net-based EDM \cite{karras2022elucidating} and the DiT-based Rectified Flow model SiT-XL/2+REPA \cite{ma2024sit,yu2024representation}. Experiments are conducted on the class-conditional ImageNet dataset \cite{deng2009imagenet} (1.28M images, 1000 classes). We report Fréchet Inception Distance (FID) \cite{heusel2017gans} computed with 50K samples as the metric for image quality.

\noindent \textbf{Implementation Details.} 
For both architectures, we employ our \textbf{Regime-Adaptive Sampling} strategy: Logit-Normal($0.8, 1.6$) for Score and Logit-Normal($-0.8, 1.6$) for Adversarial Distillation.
For ImageNet-64$\times$64, we \textbf{strictly adhere} to the DMD2 \cite{yin2024improved} configuration, modifying \textbf{only the sampling strategy} and the discriminator attachment.
Notably, despite the EDM backbone operating on a discrete and non-uniform timestep schedule which theoretically warrants specific calibration, we simply apply our sampling strategy directly over the pre-defined grid without any re-weighting or hyperparameter tuning. 
For ImageNet-256$\times$256, we distill the SiT-XL/2+REPA model using the same strategy, adapted for the native velocity space.
\textbf{More details} are provided in \textbf{Appendix~\cref{app:implementation}}.

\begin{table*}[t]
    \centering
    \begin{minipage}[t]{0.67\textwidth}
        \centering
        \caption{\textbf{Class-Conditional ImageNet 256 $\times$ 256.} Comparison with state-of-the-art fast flow models.}
        \label{tab:imagenet_256}
        \resizebox{\textwidth}{!}{
        \begin{tabular}{lcccc}
            \toprule
            \textbf{Method} & \textbf{Epochs} & \textbf{\#Params} & \textbf{NFE} $\downarrow$ & \textbf{FID} $\downarrow$ \\
            \midrule
            \multicolumn{5}{l}{\textit{Teacher Diffusion / Flow Models}} \\
            \midrule
            SiT-XL/2+REPA \cite{yu2024representation} & 800 & 675M & $250\times2$ & 1.42 \\
            \midrule
            \multicolumn{5}{l}{\textit{Fast Flow from scratch}} \\
            \midrule
            \multirow{2}{*}{Shortcut-XL/2 \cite{frans2024one}} & \multirow{2}{*}{250} & \multirow{2}{*}{675M} & 1 & 10.60 \\
             & & & 128 & 3.80 \\
            \midrule
            \multirow{2}{*}{IMM-XL/2 \cite{zhou2025inductive}} & \multirow{2}{*}{3840} & \multirow{2}{*}{675M} & $1 \times 2$ & 7.77 \\
             & & & $8 \times 2$ & 1.99 \\
            \midrule
            \multirow{2}{*}{STEI \cite{liu2025learning}} & \multirow{2}{*}{1420} & \multirow{2}{*}{675M} & 1 & 7.12 \\
             & & & 8 & 1.96 \\
            \midrule
            \multirow{2}{*}{MeanFlow-XL/2 \cite{geng2025mean}} & 240 & \multirow{2}{*}{676M} & 1 & 3.43 \\
             & 1000 & & 2 & 2.20 \\
            \midrule
            \multirow{2}{*}{iMF-XL/2 \cite{geng2025improved}} & \multirow{2}{*}{800} & \multirow{2}{*}{610M} & 1 & 1.72 \\
            & & & 2 & 1.54 \\
            \midrule
            \midrule
            \multicolumn{5}{l}{\textit{Fast Flow by distillation}} \\
            \midrule
            \multicolumn{5}{l}{\textit{Teacher: SiT-XL/2+REPA (FID = 1.42)}} \\
            \midrule
            FACM \cite{peng2025facm} & -- & 675M & 2 & 1.52 \\
            \midrule
            \multirow{2}{*}{$\pi$-Flow \cite{chen2025pi}} & \multirow{2}{*}{448} & \multirow{2}{*}{675M} & 1 & 2.85 \\
             & & & 2 & 1.97 \\
            \midrule
            \multirow{2}{*}{DMF-XL/2+ \cite{lee2025decoupled}} & \multirow{2}{*}{80} & \multirow{2}{*}{675M} & 1 & 2.16 \\
             & & & 4 & 1.51 \\
            \midrule
            \multirow{2}{*}{FreeFlow-XL/2 \cite{tong2025flow}} & 20 & \multirow{2}{*}{678M} & \multirow{2}{*}{1} & 1.84 \\
             & 300 & & & 1.45 \\
            \midrule
            \textbf{EASD-XL/2 (Ours)} & 20 & \multirow{3}{*}{675M} & \multirow{3}{*}{1} & 1.49 \\
            \hspace{0.3cm}\textit{w/ Fake Score EMA (decay scale=0.9)} & 60 & & & \textbf{1.41} \\
            \hspace{0.3cm}\textit{w/ Iterative Warmup (5k/20K)} & 60 & & & \textbf{1.39} \\
            \bottomrule
        \end{tabular}
        }
    \end{minipage}
    \hfill
    \begin{minipage}[t]{0.313\textwidth}
        \centering
        \caption{\textbf{Class-Conditional ImageNet 64 $\times$ 64.}
        Following the protocol in DMD2 \cite{yin2024improved}, ``longer training'' denotes fine-tuning from a checkpoint pre-trained exclusively with score distillation.}
        \label{tab:imagenet_64}
        \resizebox{\textwidth}{!}{
        \begin{tabular}{lcc}
            \toprule
            \textbf{Method} &\textbf{FID} $\downarrow$ & \textbf{NFE}$\downarrow$ \\
            \midrule
            \multicolumn{3}{l}{\textit{Multi-step diffusion models}} \\
            \midrule
            EDM (Teacher) \cite{karras2022elucidating} & 2.35 & 79 \\
            RIN \cite{jabri2022scalable} & 1.23 & 1000 \\
            DisCo-Diff \cite{xu2024disco} & 1.22 & 623 \\
            \midrule
            \multicolumn{3}{l}{\textit{GANs}} \\
            \midrule
            BigGAN-deep \cite{brock2018large} & 4.06 & 1 \\
            StyleGAN-XL \cite{sauer2022stylegan} & 1.52 & 1 \\
            \midrule
            \multicolumn{3}{l}{\textit{Diffusion distillation}} \\
            \midrule
            DSNO \cite{zheng2023fast} & 7.83 & 1 \\
            iCT-deep \cite{song2023improved} & 3.25 & 1 \\
            Moment Matching \cite{salimans2024multistep} & 3.00 & 1 \\
            DMD \cite{yin2024one} & 2.62 & 1 \\
            ECM \cite{geng2024consistency} & 2.49 & 1 \\
            TCM \cite{lee2024truncated} & 2.20 & 1 \\
            EMD \cite{xie2024distillation} & 2.20 & 1 \\
            CTM \cite{kim2023consistency} & 1.92 & 1 \\
            SiD \cite{zhou2024score} & 1.52 & 1 \\
            GDD-I \cite{zheng2024diffusion} & 1.16 & 1 \\
            $f$-distill-JS \cite{xu2025one} & 1.21 & 1 \\
            $f$-distill-Forward KL & 1.16 & 1 \\
            \midrule
            \multicolumn{3}{l}{\textit{Hybrid Strategy Baselines}} \\
            \midrule
            DMD2 (200K steps) \cite{yin2024improved} & 1.51 & 1 \\
            \quad longer training (140K steps) & 1.28 & 1 \\
            \midrule
            \multicolumn{3}{l}{\textit{Ours}} \\
            \midrule
            \textbf{DMD2 + EASD (100K steps)} & \textbf{1.46} & 1 \\
            \textbf{DMD2 + EASD (200K steps)} & \textbf{1.26} & 1 \\
            \quad longer training (50K steps) & \textbf{1.27} & 1 \\
            \quad longer training (140K steps) & \textbf{1.15} & 1 \\
            \quad longer training (200K steps) & \textbf{1.13} & 1 \\
            \bottomrule
        \end{tabular}
        }
    \end{minipage}
    \vspace{-7pt}
\end{table*}

\subsection{Comparison with State-of-the-Art}

 \noindent \textbf{ImageNet-256$\times$256 Results.}
As presented in \cref{tab:imagenet_256}, EASD exhibits exceptional training efficiency on high-resolution generation. With only \textbf{20 epochs}, we achieve a competitive FID of 1.49, already outperforming prolonged training runs of other methods.
Extending training to 60 epochs yields a new state-of-the-art FID of \textbf{1.39} (using Iterative Warmup) or \textbf{1.41} (using Fake Score EMA), surpassing recent fast flow baselines.
The consistent gains across varying resolutions (64, 256), diverse backbone architectures (U-Net, DiT), and distinct generative paradigms (Score-based Diffusion, Flow Matching) strongly validate the universality and robustness of the EASD framework.

\noindent \textbf{ImageNet-64$\times$64 Results.}
As summarized in \cref{tab:imagenet_64}, EASD demonstrates superior efficiency and generation quality compared to existing baselines.
Under the standard training schedule, EASD surpasses the strong hybrid baseline DMD2 \cite{yin2024improved} with significantly fewer resources: achieving an FID of \textbf{1.46} in just 100K steps versus DMD2's 1.51 in 200K steps—a \textbf{2$\times$ convergence speedup}.
Furthermore, in the ``longer training" regime (fine-tuning from score-only checkpoints), our method pushes the performance envelope further. We achieve a remarkable FID of \textbf{1.15} at 140K steps vs. DMD2's 1.28. Furthermore, we provide in \cref{fig:FID convergence curves} the FID convergence curves of EASD and DMD2.
These results confirm that our EASD strategy not only accelerates convergence but also raises the performance ceiling of hybrid distillation by effectively allocating supervision signals.

\textbf{More experiments} on \textbf{LAION~\cite{schuhmann2021laion}} and \textbf{CIFAR-10~\cite{krizhevsky2009learning}} are provided in \textbf{Appendix~\cref{app:transferability}}. 

\subsection{Qualitative Results}
\label{subsec:more_samples}

As shown in \cref{fig:qualitative_results,fig:more_samples,fig:more_samples2}, We present samples generated by our EASD-XL/2 model trained on Class-Conditional ImageNet $256 \times 256$. 
All images are synthesized in \textbf{one inference step}.

\subsection{Ablation Studies}
\label{subsec:ablation}

We conduct comprehensive ablation studies on ImageNet-256 using the SiT-XL/2+REPA backbone.

\noindent \textbf{Decoupling Effective Regimes.}
We empirically verify the complementary roles of the two objectives by training them individually with Logit-Normal sampling biased towards different temporal regimes.
As shown in Table~\cref{tab:separate_regimes} (Left), \textbf{Score Distillation} achieves its optimal FID of 2.14 when biased towards the high-noise regime ($\mathcal{LN}(0.8, 1.6)$). Biasing it towards low noise ($\mathcal{LN}(-0.8, 1.6)$) drastically degrades performance to 2.99, confirming that score gradients are more effective for global structural formation (high $t$).
Conversely, Table~\cref{tab:separate_regimes} (Right) shows that \textbf{Adversarial Distillation} performs best in the low-noise regime ($\mathcal{LN}(-0.8, 1.6)$, FID 2.31). Biasing it towards high noise ($\mathcal{LN}(0.8, 1.6)$) degrades FID to 2.80 due to signal loss. This confirms adversarial gradients are more effective for texture (low $t$).
\begin{table*}[t]
    \centering
    \begin{minipage}[t]{0.53\textwidth}
        \centering
        \caption{\textbf{Decoupling Regimes.} FID comparison of individual objectives under different sampling biases. $\mathcal{LN}(\mu, s)$ denotes Logit-Normal sampling.}
        \label{tab:separate_regimes}
        \begin{minipage}[t]{0.49\textwidth}
            \centering
            \subcaption{\textbf{Score Distillation}}
            \resizebox{\textwidth}{!}{
                \begin{tabular}{cc}
                    \toprule
                    Sampling Bias & FID $\downarrow$ \\
                    \midrule
                    $\mathcal{LN}(-0.8, 1.6)$ & 2.99 \\
                    $\mathcal{LN}(0.0, 1.6)$  & 2.28 \\
                    $\mathcal{LN}(0.8, 1.6)$  & \textbf{2.14} \\
                    $\mathcal{LN}(0.8, 1.2)$  & 2.16 \\
                    $\mathcal{LN}(0.8, 2.0)$  & 2.31 \\
                    \bottomrule
                \end{tabular}
            }
        \end{minipage}
        \hfill
        \begin{minipage}[t]{0.49\textwidth}
            \centering
            \subcaption{\textbf{Adversarial Distillation}}
            \resizebox{\linewidth}{!}{
                \begin{tabular}{cc}
                    \toprule
                    Sampling Bias & FID $\downarrow$ \\
                    \midrule
                    $\mathcal{LN}(0.8, 1.6)$  & 2.80 \\
                    $\mathcal{LN}(0.0, 1.6)$  & 2.54 \\
                    $\mathcal{LN}(-0.8, 1.6)$ & \textbf{2.31} \\
                    $\mathcal{LN}(-0.8, 1.2)$ & Fail \\
                    $\mathcal{LN}(-0.8, 2.0)$ & 2.37 \\
                    \bottomrule
                \end{tabular}
            }
        \end{minipage}
    \end{minipage}
    \hfill
    \begin{minipage}[t]{0.44\textwidth}
        \centering
        \vspace{-\abovecaptionskip} 
        \caption{\textbf{Synergy Analysis.} Performance comparison of different sampling strategy combinations. We evaluate the baseline Uniform sampling ($\mathcal{U}$), Hard Truncation, and Logit-Normal ($\mathcal{LN}$) strategy.}
        \label{tab:ablation_sampling}
        \vspace{7pt} 
        \resizebox{\textwidth}{!}{
            \begin{tabular}{ccc}
                \toprule
                Score Sampling & Adversarial Sampling & FID $\downarrow$ \\
                \midrule
                $\mathcal{U}(0.0, 1.0)$   & $\mathcal{U}(0.0, 1.0)$   & 1.60 \\
                $\mathcal{U}(0.2, 1.0)$   & $\mathcal{U}(0.0, 0.8)$   & 1.51 \\
                $\mathcal{LN}(0.8, 1.6)$  & $\mathcal{U}(0.0, 0.8)$   & 1.50 \\
                $\mathcal{U}(0.2, 1.0)$   & $\mathcal{LN}(-0.8, 1.6)$ & 1.49 \\
                $\mathcal{LN}(0.8, 1.6)$  & $\mathcal{LN}(-0.8, 1.6)$ & \textbf{1.47} \\
                \bottomrule
            \end{tabular}
        }
    \end{minipage}
    \vspace{-14pt}
\end{table*}

\begin{table}[t]
\centering
\caption{\textbf{Strict Ablation with DMD2 on ImageNet-$64\times 64$} (5:1 update ratio, batch size 280; all runs update Gen. + Fake Score in score distillation). $^*$Fake-score is jointly updated in discriminator optimization. $^\dagger$Frozen-teacher discriminator without updating fake-score.}
\label{tab:strict_dmd2}
\footnotesize
\setlength{\tabcolsep}{10pt}
\begin{tabular}{lccccc}
\toprule
\textbf{Method} & \textbf{Adversarial Updates} & \textbf{Time} & \textbf{FID@100K $\downarrow$} & \textbf{FID@200K $\downarrow$} \\
\midrule
DMD2~\cite{yin2024improved} & Gen. + Disc. + Fake Score$^*$ & 28h & 1.72 & 1.52 \\
+ Regime-Adaptive Sampling & Gen. + Disc. + Fake Score$^*$ & 28h & 1.50 & 1.35 \\
\textbf{EASD (Full)} & Gen. + Disc.$^\dagger$ & \textbf{23h} & \textbf{1.46} & \textbf{1.26} \\
\bottomrule
\end{tabular}
\vspace{-10pt}
\end{table}

\noindent \textbf{Optimal Synergy Strategy.}
Table~\ref{tab:ablation_sampling} \textbf{validates our guideline: encouraging each distillation objective to operate more within its complementary effective regime consistently improves hybrid distillation performance.} The baseline using standard Uniform sampling for both objectives yields a suboptimal FID of 1.60. Simply applying Hard Truncation to bias Score Distillation toward high-noise timesteps and Adversarial Distillation toward low-noise timesteps already improves FID to 1.51, confirming that the principle holds regardless of the specific instantiation. Replacing either objective's sampling with a Logit-Normal distribution further improves performance (FID 1.50 and 1.49), and combining both yields the best result of \textbf{1.47} with our proposed Regime-Adaptive Sampling Strategy. This demonstrates that soft biasing via Logit-Normal distributions, which maintains full temporal support while concentrating supervision on effective regimes, offers superior synergy over hard truncation. More combination strategies and transferability experiments are provided in Appendix~\cref{app:ablation}.

\noindent \textbf{Discriminator Architecture Design.}
To validate our discriminator design, we conduct experiments using the optimal sampling strategy (trained for 40 epochs). As shown in Table~\cref{tab:ablation_components} (Left), attaching the discriminator heads to the \textbf{Frozen Teacher} significantly outperforms attaching them to the updating Fake Score network (FID 1.47 vs. 1.85), confirming that non-stationary student features destabilize adversarial training.
We also investigate the optimal attachment depth within the 28-block DiT backbone. Connecting the discriminator head to the middle block (Block 14, default setting) outperforms connecting to shallow (Block 7, FID 1.55) or deep (Block 21, FID 1.70) blocks. This suggests that middle block offer better trade-off between semantic and low-level feature representation. Additionally, projecting DiT tokens to 2D grids for standard CNN discriminators yields a suboptimal FID of 1.55, demonstrating the efficiency of our design. \textbf{More attempts} are in \textbf{Appendix~\cref{app:negative_results}}.

\noindent \textbf{Stabilization and Quality Refinement.}
To strictly isolate the impact of our stabilization techniques, we employ the same configuration as the first row of \cref{tab:ablation_sampling}: a baseline using full-range Uniform sampling for both distillation objectives. We observe that this naive setup is inefficient and unstable, with performance saturating early at an FID of 1.60 after just 20 epochs.
Introducing \textbf{Fake Score EMA} breaks this bottleneck, stabilizing the adversarial dynamics and improving FID to 1.56. Alternatively, employing \textbf{Warmup} (first 5K steps) achieves a comparable effect (FID 1.57), confirming that damping initial oscillations is useful.
Additionally, incorporating the \textbf{Latent Mode Seeking (LMS)} loss improves the baseline to 1.58, demonstrating its ability to restore diversity.
Finally, we note that removing the Student Model EMA causes catastrophic degradation (FID 1.90), underscoring that parameter averaging is a fundamental prerequisite for distilling DiT-based models.

\textbf{Strict Comparison under the Same Hybrid Paradigm (DMD2).} 
Both DMD2~\cite{yin2024improved} and EASD adopt the same hybrid distillation paradigm combining score and adversarial objectives. To strictly isolate the gain of regime-adaptive synergy over a naive combination, we conduct a controlled ablation directly on the official DMD2 codebase (ImageNet-$64\times 64$) by modifying \textit{only} the timestep sampling strategy while keeping all other settings (learning rate, 5:1 update schedule, batch size 280, and architecture) identical. As shown in Table~\ref{tab:strict_dmd2}, simply reallocating timestep regimes via our Regime-Adaptive Sampling within the identical hybrid framework improves FID from 1.72 to 1.50 at 100K steps and from 1.52 to 1.35 at 200K steps under the exact same training time (28h), matching DMD2's 200K-step performance in just 100K steps. Equipping the model with our Frozen-Teacher Discriminator in full EASD further avoids updating the fake-score backbone during adversarial steps, reducing training time to 23h while pushing FID to 1.46 (100K) and 1.26 (200K).

\section*{Acknowledgements}
This work was supported by the STI 2030-Major Projects under Grant 2021ZD0201404 and Shenzhen Key Laboratory of New Generation Interactive Media Technology Innovation(ZDSYS20210623092001004).

\bibliographystyle{plainnat}
\bibliography{reference}
\newpage
\appendix
\onecolumn

\section{Limitations}
\label{app:limitation}
In this work, we analyze the duality between adversarial distillation and score distillation, along with their respective independent shortcomings. We investigate the prerequisites for combining the two and the potential mechanisms through which their integration can synergistically enhance diffusion distillation. We characterize the complementarity of score distillation and adversarial distillation across different timestep regimes, identify the inefficiency inherent in the current naive hybrid paradigm, and propose Regime-Adaptive Sampling to further improve the combinatorial efficiency of hybrid distillation, achieving faster training convergence and better final performance. 

Nevertheless, the following limitations remain in this work. First, the proposed method is primarily oriented toward hybrid distillation frameworks. While it also provides certain benefits for single-objective distillation, the inherent and unavoidable deficiencies of single-objective distillation suggest that introducing additional supervisory signals is still advisable in such settings. Second, although the proposed guidance methodology effectively improves distillation in our experiments, and we demonstrate that any optimization strategy aligned with the core principle—encouraging different distillation objectives to operate within each other's complementary effective timestep regions—can improve hybrid distillation performance, the specific sampling distributions involve hyperparameter configurations that may require further tuning when directly transferred to other datasets or frameworks to achieve optimal results. Nonetheless, our proposed configurations exhibit strong transferability across settings. Third, our assessment of distillation effectiveness relies primarily on the widely adopted FID metric. For single-step distillation in particular, models with identical FID scores may still exhibit inferior perceptual quality compared to multi-step models from a human visual perspective, and future work should incorporate human preference evaluation metrics to provide a more comprehensive assessment.

\section{More Quantitative Experiments}

\subsection{More Transferability Experiments}
\label{app:transferability}

We apply EASD to the text-to-image model SDXL \cite{podell2023sdxl} using the same experimental setup as DMD2 \cite{yin2024improved}, adjusting only the sampling strategy. As mentioned in \cref{subsec:latent_regularization}, for the t2i task, we choose to retain CFG=8 for distillation. The model is trained on the LAION  \cite{schuhmann2021laion} dataset, and we report results at 18K training steps, fully inheriting the hyperparameter settings from the ImageNet experiments (without any additional tuning of the logitnorm parameters). As shown in the \cref{tab:sdxl_coco}, in the 1-step generation evaluation on COCO-10K \cite{lin2014microsoft}, EASD surpasses the hybrid distillation method DMD2 (25K steps) and ADD \cite{wei2024adversarial} with fewer training steps. 

We train a UNet-based flow matching model from scratch on CIFAR-10 \cite{krizhevsky2009learning} for 2000 epochs using linear timestep sampling. The FID evaluated with 100-step Euler inference is 3.27. We conduct an ablation study on score distillation and adversarial distillation sampling strategies on this model, training each configuration for 100 epochs. The single-step evaluation results shown in \cref{tab:cifar10_ablation}.

\begin{table}[h]
\centering
\caption{Comparison of 1-step generation on COCO-10K with SDXL-based methods.}
\label{tab:sdxl_coco}
\begin{tabular}{lcc}
\toprule
\textbf{Method (1-step SDXL)} & \textbf{COCO-10K FID $\downarrow$} & \textbf{CLIP Score $\uparrow$} \\
\midrule
SDXL-Turbo (ADD) & 23.92 & 0.337 \\
DMD2             & 19.01 & 0.336 \\
\textbf{EASD (Ours)} & \textbf{18.81} & \textbf{0.341} \\
\bottomrule
\end{tabular}
\end{table}

\begin{table}[h]
\centering
\caption{Ablation study of sampling strategy combinations for score distillation and adversarial distillation on CIFAR-10.}
\label{tab:cifar10_ablation}
\begin{tabular}{ccc}
\toprule
\textbf{Score Distillation} & \textbf{Adversarial Distillation} & \textbf{FID $\downarrow$} \\
\midrule
$\mathcal{U}(0.0, 1.0)$       & $\mathcal{U}(0.0, 1.0)$        & 6.98 \\
$\mathcal{LN}(0.8, 1.6)$      & $\mathcal{U}(0.0, 1.0)$        & 5.58 \\
$\mathcal{U}(0.0, 1.0)$       & $\mathcal{LN}(-0.8, 1.6)$      & 6.06 \\
$\mathcal{LN}(1.0, 1.6)$      & $\mathcal{LN}(-1.0, 1.6)$      & 5.11 \\
$\mathcal{LN}(0.8, 1.6)$      & $\mathcal{LN}(-0.8, 1.6)$      & \textbf{5.02} \\
$\mathcal{LN}(0.4, 1.6)$      & $\mathcal{LN}(-0.4, 1.6)$      & 5.61 \\
$\mathcal{LN}(0.0, 1.6)$      & $\mathcal{LN}(0.0, 1.6)$       & 6.97 \\
$\mathcal{LN}(0.8, 1.0)$      & $\mathcal{LN}(-0.8, 1.0)$      & 7.57 \\
$\mathcal{LN}(0.8, 2.0)$      & $\mathcal{LN}(-0.8, 2.0)$      & 5.09 \\
\bottomrule
\end{tabular}
\end{table}

The results demonstrate that the hyperparameters established in our main experiments exhibit good transferability. Furthermore, configurations in the vicinity of our chosen parameters yield comparable performance, which is consistent with the core motivation of this work: encouraging the two distillation objectives to operate more within each other's effective timestep regions—score distillation favoring high-noise timesteps and adversarial distillation favoring low-noise timesteps, which are precisely the ineffective regions for the respective counterpart—can improve the overall efficiency of hybrid distillation. Achieving optimal performance requires more fine-grained tuning and a broader search over candidate configurations.

\subsection{More Efficiency Experiments}
\label{app:efficiency}

In Table~\ref{tab:efficiency}, we compare against state-of-the-art distillation methods \cite{lee2025decoupled, chen2025pi}  by distilling the identical base model (SiT-XL/2+ \cite{yu2024representation}) on the same GPU, reporting key metrics under each method's official configuration. All experiments are trained for 20 epochs on the same amount of data. We provide a detailed comparison table of key experimental metrics:

\begin{table}[h]
\centering
\caption{Efficiency comparison of distillation methods on SiT-XL/2+REPA \cite{yu2024representation}.}
\label{tab:efficiency}
\resizebox{\textwidth}{!}{%
\begin{tabular}{lcccccc}
\toprule
\textbf{Method} & \textbf{Memory (BS=256)} & \textbf{Best-Practice Config \& Memory} & \textbf{20-Epoch Time} & \textbf{FID (20ep) $\downarrow$} & \textbf{Best-Practice Time} & \textbf{Best FID $\downarrow$} \\
\midrule
PiFlow \cite{chen2025pi} & 17.8 GB & BS=4096, 35.8 GB & 6h 10min  & 5.13 & 138h & 2.85 \\
DMF  \cite{lee2025decoupled}    & 56.4 GB & BS=256,  56.4 GB & 11h 30min & 2.79 & 46h  & 2.16 \\
EASD (Ours) & 48.7 GB & BS=256, 48.7 GB & 8h 30min & 1.47 & 26h & 1.39 \\
\bottomrule
\end{tabular}%
}
\end{table}

PiFlow \cite{chen2025pi} accelerates training by appending additional layers to the network, resulting in lower GPU memory usage at the same batch size. However, its official documentation explicitly states that a large batch size of 4096 is required for effective training, and its convergence is relatively slow, requiring 138 hours to reach the best reported result.

DMF \cite{lee2025decoupled} distills the model using a consistency model formulation (also adopted by FACM \cite{peng2025facm} and FreeFlow \cite{tong2025flow}), requiring computationally expensive Jacobian-vector product (JVP) calculations to update the model. This leads to high memory consumption and slow training throughput. Despite leveraging the advanced FlashAttention-3 \cite{shah2024flashattention} in the official codebase, significant overhead remains.

Our EASD achieves substantially faster convergence, shorter total training time, and moderate memory consumption.

\begin{figure}
    \centering
    \includegraphics[width=0.75\linewidth]{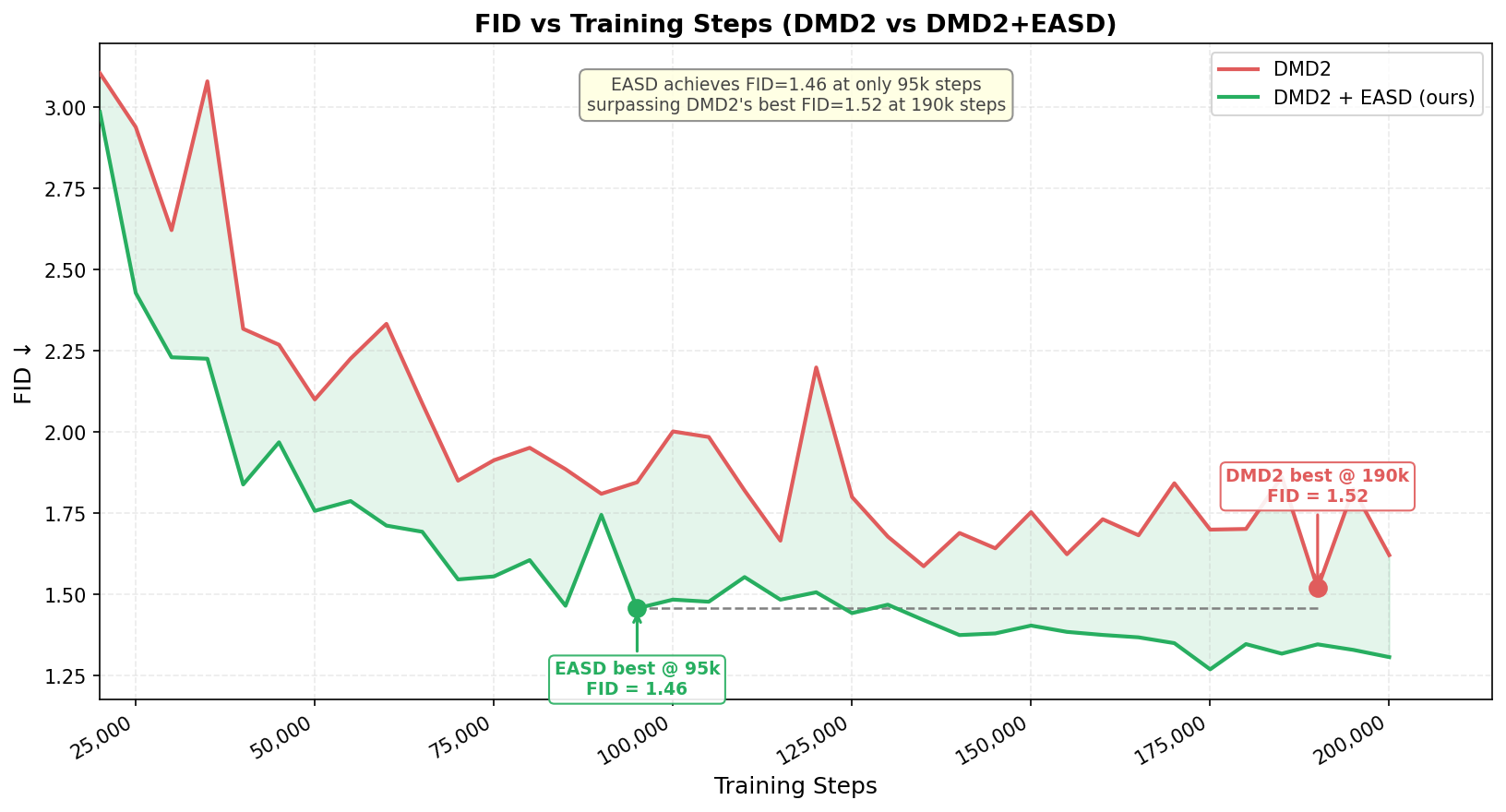}
    \caption{FID convergence curves.}
    \label{fig:FID convergence curves}
\end{figure}

Furthermore, we provide in \cref{fig:FID convergence curves} the FID convergence curves of EASD and DMD2 on EDM~\cite{karras2022elucidating}, corresponding to the comparison reported in \cref{tab:imagenet_64}, demonstrating that our method achieves significant acceleration in convergence efficiency.

\subsection{More Ablation Studies on Sampling Strategies}
\label{app:ablation}

We additionally evaluate three combinations involving truncated log-normal (LN) distributions:

\begin{table}[h]
\centering
\caption{Ablation study on truncated sampling strategies for score distillation and adversarial distillation.}
\label{tab:truncated_ablation}
\begin{tabular}{llc}
\toprule
\textbf{Score Distillation} & \textbf{Adversarial Distillation} & \textbf{FID $\downarrow$} \\
\midrule
$\mathcal{LN}(0.8, 1.6)$, truncated to $(0,\ 0.2)$ & $\mathcal{LN}(-0.8, 1.6)$, truncated to $(0.8,\ 1.0)$ & 1.47 \\
$\mathcal{LN}(0.8, 1.6)$, truncated to $(0,\ 0.2)$ & $\mathcal{LN}(-0.8, 1.6)$, untruncated                & 1.50 \\
$\mathcal{LN}(0.8, 1.6)$, untruncated              & $\mathcal{LN}(-0.8, 1.6)$, truncated to $(0.8,\ 1.0)$ & 1.49 \\
\bottomrule
\end{tabular}
\end{table}

The results show that, compared to the baseline (FID 1.46), additional truncation does not yield further improvement. This suggests that, while it is crucial for each distillation objective to operate within its effective timestep region, the ``ineffective'' region still provides a certain degree of regularization. Completely truncating this region eliminates such benefits. 

\section{Detailed Implementation}
\label{app:implementation}

\subsection{ImageNet-64$\times$64 (EDM Backbone)}
\label{app:impl_edm}

\textbf{Configuration and Optimization.}
We adopt the U-Net-based EDM \cite{karras2022elucidating} architecture, strictly following the DMD2 \cite{yin2024improved} setup for fair comparison. All networks (Generator, Fake Score, Discriminator) are trained using AdamW with a fixed learning rate of $2 \times 10^{-6}$, $\beta_1=0.9$, $\beta_2=0.999$, weight decay of $10^{-2}$, and a batch size of 280. We employ a TTUR strategy \cite{heusel2017gans}, updating the critics (Discriminator and Fake Score) 5 times per generator update. A linear warmup of 500 steps is applied. For Adversarial Distillation, loss weights are set to $10^{-2}$ for the discriminator update and $3 \times 10^{-3}$ for the generator update. Consistent with our "Frozen Teacher" design, we attach the CNN-based discriminator to the encoder of the \textit{frozen real score network} rather than the updating fake score network used in DMD2.

\textbf{Discrete Timestep Sampling.}
Despite EDM's use of a discrete and non-uniform timestep schedule, we verify robustness by directly mapping our continuous Logit-Normal distribution to the EDM grid without re-calibration. Specifically, we sample $y \sim \mathcal{N}(\mu, s^2)$, compute the probability $p = \sigma(y)$, and map it to the discrete index via $t_{\text{idx}} = \lfloor N_{\min} + p \cdot (N_{\max} - N_{\min}) \rfloor$. This demonstrates that our Regime-Adaptive Sampling is effective even without tailoring to specific noise density schedules.

\textbf{Longer Training Setup.}
Following DMD2, we fine-tune a checkpoint pre-trained for 400K steps (using only Score Distillation and Regression) by removing the regression loss and adding the adversarial objective. The learning rate is reduced to $5 \times 10^{-7}$. While the original DMD2 typically converges around 140K steps, our method shows stable improvements up to 200K steps. Notably, advanced stabilization techniques (EMA, LMS) were \textit{not} employed here, isolating the gain to our sampling strategy.

\subsection{ImageNet-256$\times$256 (SiT Backbone)}
\label{app:impl_sit}

\textbf{Configuration and Optimization.}
We utilize the SiT-XL/2 \cite{ma2024sit} model initialized with REPA \cite{yu2024representation} weights, trained for 60 epochs (approx. 300K steps) with a batch size of 256. We use distinct learning rates: $3 \times 10^{-3}$ for the generator and $1 \times 10^{-4}$ for the critics (Fake Score and Discriminator), maintaining the 5:1 TTUR ratio. All networks (Generator, Fake Score, Discriminator) are trained using AdamW with $\beta_1=0.9$, $\beta_2=0.999$, and weight decay of $10^{-2}$. Adversarial loss weights match the EDM setup ($10^{-2}$ for discriminator, $3 \times 10^{-3}$ for generator).

\textbf{Sampling and Stabilization.}
We apply Regime-Adaptive Sampling with $\mathcal{LN}(0.8, 1.6)$ for Score Distillation and $\mathcal{LN}(-0.8, 1.6)$ for Adversarial Distillation.
For stabilization, we employ either \textbf{Iterative Warmup} (freezing the generator to update the Fake Score network for 5K steps every 20K cycles) or \textbf{Fake Score EMA} (decay 0.9).
Empirically, we find that these techniques serve overlapping roles in damping oscillations and yield suboptimal results when combined. Consequently, we treat them as \textbf{mutually exclusive} alternatives (i.e., enabling one disables the other). The generator consistently use a standard EMA decay of 0.999. LMS loss weight is $3 \times 10^{-3}$.

\section{The Compounding Mechanism of CFG in Score Distillation}
\label{app:recursive_guidance}

In this section, we analyze the mechanism of Classifier-Free Guidance (CFG) in Score Distillation and explain why it leads to inevitable mode collapse.

\subsection{Standard Distillation Setup}
In a standard Score Distillation setting with CFG, the objective aligns the student's output with the teacher's guided prediction.
Let the teacher's guided velocity be:
\begin{equation}
    v_{\text{teacher}}^{\text{cfg}}(x_t) = v_{\text{uncond}}(x_t) + w \cdot (v_{\text{cond}}(x_t) - v_{\text{uncond}}(x_t)).
\end{equation}
The student generator $G_\theta$, which does not use CFG during inference, is trained to minimize the difference between its output $v_{\text{student}}(x_t)$ and $v_{\text{teacher}}^{\text{cfg}}(x_t)$.
If training is effective, the student implicitly internalizes the guidance scale $w$, producing samples $x_{\text{student}}$ that exhibit the characteristics of a guided distribution (e.g., higher contrast, centered modes).

\subsection{The Recursive Feedback Loop}
The critical issue arises from the input to the teacher model during training. Unlike standard sampling where the model iterates on its own trajectory, in distillation, the teacher computes the score target based on the \textbf{student's generated sample} $\hat{x}_0$ (perturbed to $\hat{x}_t$).

Consider a converged state where the student has already learned to produce samples equivalent to the teacher with guidance scale $w$. 
When this ``guided" sample $x_{\text{student}}$ is fed back into the teacher:
1.  The teacher perceives $x_{\text{student}}$ as a sample from a distribution that has already been sharpened by $w$.
2.  The teacher then applies its CFG formula \textit{again} on this input to compute the new target.

Mathematically, if we approximate the effect of CFG as scaling the score (or velocity) vector towards the mode, applying CFG again on a sample that is already shifted towards the mode results in a multiplicative compounding of the guidance strength:
\begin{equation}
    \text{Effective Scale} \approx w_{\text{learned}} \times w_{\text{teacher}}.
\end{equation}
This creates a \textbf{recursive feedback loop}:
 \begin{enumerate}[nosep] 
    \item Student learns guidance $w$.
    \item Teacher sees sample with $w$, applies $w$ again $\to$ target effectively has $w^2$.
    \item Student updates to match $w^2$.
    \item Teacher sees sample with $w^2$, applies $w$ again $\to$ target effectively has $w^3$...
\end{enumerate}
This runaway process continuously pushes the student's distribution into narrower and narrower probability peaks. Consequently, the generator quickly collapses to producing only the single most likely sample (the mode) for each class, resulting in severe mode collapse and oversaturated artifacts.
While auxiliary losses (e.g., regression or adversarial) can mitigate this by pulling the distribution back, they do so by contradicting the score gradient, effectively sacrificing the precision of the supervision signal.

\subsection{Equilibrium at Excessive Guidance}
The recursive compounding described above does not imply an infinite explosion. In practice, the distillation process eventually reaches a stationary point, but at a \textbf{distorted equilibrium}.
As the student generator updates, the \textit{Fake Score} network (which estimates the student's score) also adapts to this increasingly sharpened distribution. The recursive stacking halts when the student successfully mimics the teacher's compounded output—that is, when the fake score $s_{\text{fake}}$ aligns with the teacher's guided score $s_{\text{teacher}}^{\text{cfg}}$ on the student's current samples.
At this equilibrium, the gradient of the distillation loss approaches zero:
\begin{equation}
    \nabla \mathcal{L} \propto (s_{\text{teacher}}^{\text{cfg}}(\hat{x}) - s_{\text{fake}}(\hat{x})) \to 0.
\end{equation}
While the superposition of CFG stops increasing, the damage is done. The system stabilizes at an \textbf{effective guidance scale} $w_{\text{eff}}$ that is larger than the intended hyperparameter $w$ (i.e., $w_{\text{eff}} \gg w$).
Consequently, the converged student produces samples that exhibit the characteristics of \textit{excessive} guidance—oversaturated colors and reduced diversity (mode collapse)—not because the math exploded to infinity, but because the model converged to an over-guided target.

\begin{figure*}[t]
    \centering
    \includegraphics[width=1.0\linewidth]{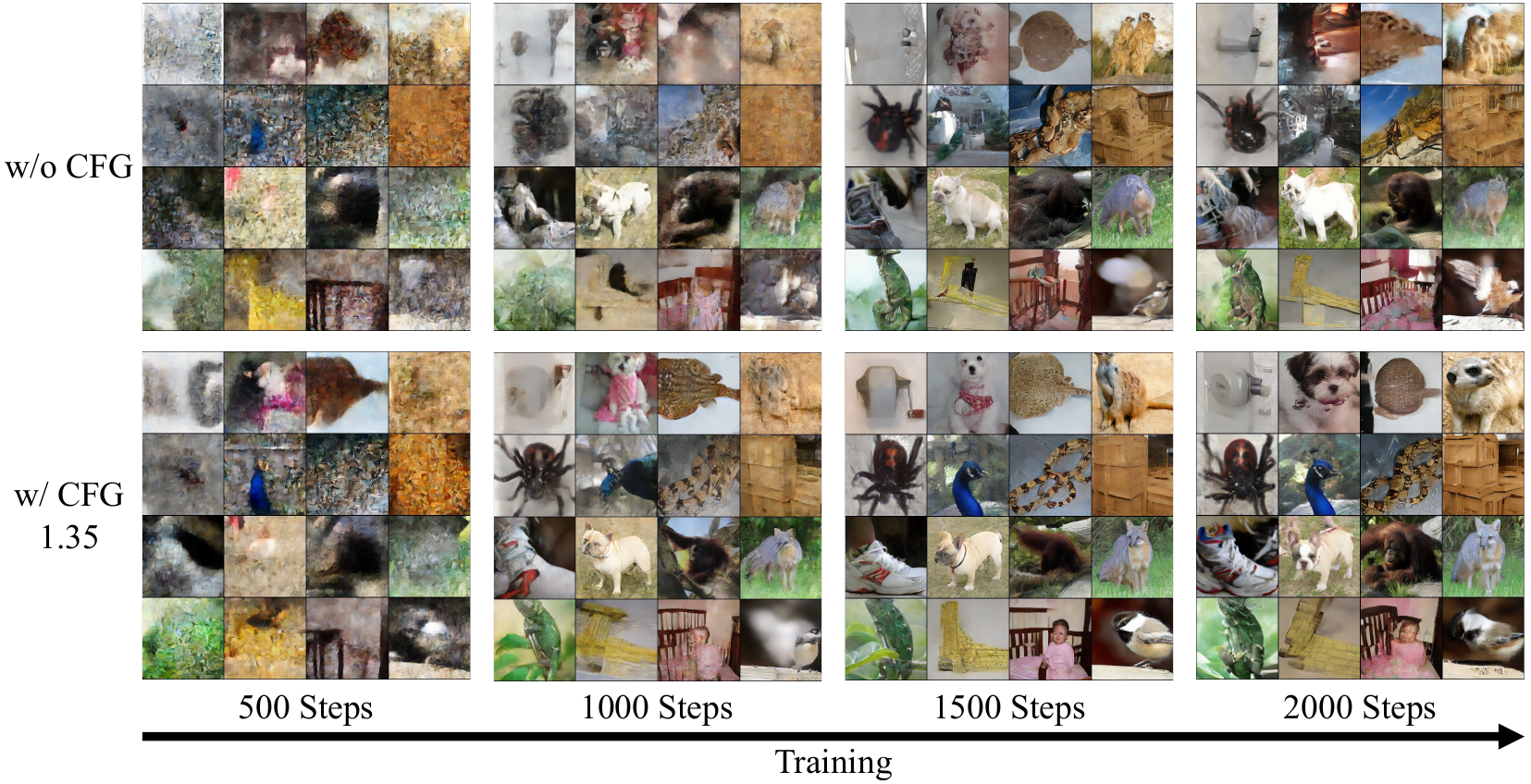}
    \caption{\textbf{Visual comparison of training dynamics with and without CFG.} 
    Rows correspond to samples generated by models trained \textbf{without CFG} (Top) and \textbf{with CFG $w=1.35$} (Bottom) across different training steps (500 to 2000).
    The model utilizing CFG demonstrates rapid initial convergence, producing recognizable objects with sharp features. However, this acceleration leads to premature convergence, where the model collapses into a limited set of high-contrast modes. In contrast, the model without CFG evolves more gradually but preserves greater diversity, avoiding the oversaturated artifacts associated with recursive guidance.}
    \label{fig:cfg_vis}
\end{figure*}

\begin{table}[t]
    \centering
    \caption{\textbf{Quantitative Impact of CFG.} Comparison of metrics for models trained with and without CFG. The use of CFG leads to higher Precision and Inception Score (IS) but significantly lower Recall and worse FID, indicating severe mode collapse.}
    \label{tab:cfg_metrics}
    \setlength{\tabcolsep}{5pt}
    \begin{tabular}{lccccc}
        \toprule
        Method & IS $\uparrow$ & FID $\downarrow$ & sFID $\downarrow$ & Precision $\uparrow$ & Recall $\uparrow$ \\
        \midrule
        w/o CFG & 233.32 & \textbf{2.30} & \textbf{4.67} & 0.787 & \textbf{0.624} \\
        w/ CFG ($w=1.35$)       & \textbf{341.98} & 3.92 & 5.05 & \textbf{0.875} & 0.508 \\
        \bottomrule
    \end{tabular}
\end{table}

\subsection{Empirical Validation}
\label{app:cfg_experiment}

To empirically validate the theoretical analysis of recursive guidance, we conduct an ablation study comparing distillation with and without Classifier-Free Guidance (CFG). We apply a moderate guidance scale of $w=1.35$ during the distillation of the SiT-XL/2+REPA model on ImageNet-256.

\textbf{Qualitative Analysis (Visual Convergence).}
As visualized in\cref{fig:cfg_vis}, the model trained \textbf{with CFG} ($w=1.35$) exhibits extremely rapid convergence. Even at early stages (e.g., 500 steps), it generates images with distinct class features and sharp contrasts. In contrast, the model \textbf{without CFG} produces blurrier, less distinct outputs in the early phase.
However, this initial "sharpness" comes at a cost. The CFG-trained model quickly locks into specific high-likelihood patterns, creating images that appear "hyper-real" or oversaturated, a hallmark of the recursive feedback loop described in \cref{app:recursive_guidance}.

\textbf{Quantitative Analysis (Mode Collapse).}
\cref{tab:cfg_metrics} presents the quantitative metrics. The results reveal a stark trade-off:
\begin{itemize}[leftmargin=*, itemsep=2pt, topsep=0pt, parsep=2.5pt]
    \item \textbf{High Precision, Low Recall:} The CFG model achieves significantly higher Inception Score (342.0 vs. 233.3) and Precision (0.88 vs. 0.79), indicating that the generated samples are individually high-quality and belong to the target class. However, the Recall is drastically lower (0.51 vs. 0.62), proving that the model covers a much smaller portion of the data distribution (i.e., Mode Collapse).
    \item \textbf{FID Degradation:} Despite the higher visual sharpness, the FID of the CFG model is worse (3.92) than the baseline (2.30). This confirms that the loss in diversity (Recall) outweighs the gain in individual sample quality.
\end{itemize}
These empirical findings strongly support our decision to omit CFG in the default training setup, prioritizing distribution coverage (Recall) over initial visual sharpness.

\section{Detailed Derivation of Adversarial Distillation as Score Matching}
\label{app:gan_score_derivation}

In this section, we provide a rigorous derivation of the relationship between the Adversarial Distillation objective and Score Matching, as discussed in \cref{subsec:duality}. We further analyze the approximation error introduced by the scalar weighting term in the Non-Saturating GAN loss.

\subsection{Gradient of the Non-Saturating GAN Loss}

Consider the generator's objective in a standard Non-Saturating GAN (NS-GAN) framework, which aims to minimize the negative log-likelihood of the discriminator identifying generated samples as real:
\begin{equation}
    \mathcal{L}_{G}^{\text{adv}}(\theta) = - \log D_\psi(x),
\end{equation}
where $x = G_\theta(z)$ represents a generated sample. The discriminator $D_\psi(x)$ is typically parameterized as a sigmoid function acting on a logit $f_\psi(x)$:
\begin{equation}
    D_\psi(x) = \sigma(f_\psi(x)) = \frac{1}{1 + e^{-f_\psi(x)}}.
    \label{eq:sig}
\end{equation}
To analyze the update direction provided to the generator, we compute the gradient of $\mathcal{L}_{\text{GAN}}$ with respect to the sample $x$. Applying the chain rule:
\begin{equation}
    \nabla_x \mathcal{L}_{G}^{\text{adv}}(\theta) = \nabla_x \left( - \log D_\psi(x) \right) = - \frac{1}{D_\psi(x)} \nabla_x D_\psi(x).
\end{equation}
Using the derivative property of the sigmoid function, $\sigma'(u) = \sigma(u)(1 - \sigma(u))$, we have:

\begin{equation}
    \begin{aligned}
        \nabla_x \mathcal{L}_{G}^{\text{adv}}(\theta) &= - \frac{1}{D_\psi(x)} \cdot \left[ D_\psi(x) (1 - D_\psi(x)) \nabla_x f_\psi(x) \right] \\
        &= - (1 - D_\psi(x)) \nabla_x f_\psi(x).
    \end{aligned}
    \label{eq:exact_gan_gradient}
\end{equation}

\subsection{Connection to Score Difference}

We now relate the logit gradient $\nabla_x f_\psi(x)$ to the score functions. To do this, we first establish the algebraic form of the logit and then substitute the optimal discriminator solution.

\textbf{Step 1: The Logit as Log-Odds.}
By definition, the discriminator output $D(x)$ is the sigmoid activation of the logit $f(x)$ (\cref{eq:sig}).

Solving this equation for $f(x)$ allows us to express the logit in terms of the probability $D(x)$.
Taking the natural logarithm yields the log-odds form:
\begin{equation}
    f(x) = - \log \left( \frac{1 - D(x)}{D(x)} \right) = \log \left( \frac{D(x)}{1 - D(x)} \right).
    \label{eq:logit_inverse}
\end{equation}

\textbf{Step 2: Substitution of the Optimal Discriminator.}
For fixed data and model distributions, the optimal discriminator $D^*(x)$ is given by $D^*(x) = \frac{p_{\text{data}}(x)}{p_{\text{data}}(x) + p_\theta(x)}$ \cite{goodfellow2014generative}. 
Substituting this into the ratio term in \cref{eq:logit_inverse}, the expression simplifies directly to the likelihood ratio:
\begin{equation}
    \frac{D^*(x)}{1 - D^*(x)} = \frac{p_{\text{data}}(x)}{p_\theta(x)}.
\end{equation}
Consequently, the optimal logit is simply the difference of the log-densities:
\begin{equation}
    f^*(x) = \log \left( \frac{p_{\text{data}}(x)}{p_\theta(x)} \right) = \log p_{\text{data}}(x) - \log p_\theta(x).
\end{equation}

\textbf{Step 3: Recovering the Score Difference.}
Finally, taking the gradient with respect to the input sample $x$, we obtain the relationship between the logit gradient and the score functions:
\begin{equation}
    \nabla_x f^*(x) = \nabla_x \log p_{\text{data}}(x) - \nabla_x \log p_\theta(x).
    \label{eq:logit_score_final}
\end{equation}
This derivation proves that the logit gradient of an optimal GAN discriminator explicitly estimates the difference between the data score and the model score. Combining this with the result from \cref{eq:exact_gan_gradient}, the total update direction for the generator is:
\begin{equation}
    \nabla_x \mathcal{L}_{G}^{\text{adv}}(\theta) = - (1 - D^*(x)) \left( \nabla_x \log p_{\text{data}}(x) - \nabla_x \log p_\theta(x) \right).
    \label{eq:weighted_score_diff}
\end{equation}

\subsection{Analysis of the Approximation}

Equation \eqref{eq:weighted_score_diff} reveals that the GAN gradient is a vector field aligned with the score difference, scaled by a position-dependent weight $w(x) = 1 - D^*(x)$.

\textbf{Directional Consistency.}
Crucially, the scalar weight $w(x) \in (0, 1)$ is strictly positive. Therefore, the update direction $-\nabla_x \mathcal{L}_{G}^{\text{adv}}(x)$ is always positively aligned with the score difference vector $\nabla_x \log p_{\text{data}}(x) - \nabla_x \log p_\theta(x)$. This confirms the claim in the main text that the adversarial objective fundamentally drives samples towards the high-density regions of the data distribution, functionally similar to Score Distillation Sampling (SDS).

\textbf{Role of the Weight Term.}
The weight term can be expanded as:
\begin{equation}
    1 - D^*(x) = \frac{p_\theta(x)}{p_{\text{data}}(x) + p_\theta(x)}.
\end{equation}
This scaling factor provides an adaptive mechanism:
\begin{itemize}
    \item When samples are clearly unrealistic ($p_\theta \ll p_{\text{data}}$), $D^* \to 1$ and the gradient vanishes, preventing unstable updates from outliers.
    \item When samples are in ambiguous regions or overlapping with data modes, the weight remains significant, ensuring continued alignment.
\end{itemize}
In the context of our proposed method, this adaptive scaling complements the standard Score Distillation, particularly in the low-noise regime where fine-grained discrimination is necessary. By approximating $\nabla_x \mathcal{L}_{G}^{\text{adv}} \approx \nabla_x f^*(x)$ in our qualitative analysis, we focus on the geometric direction of the update, which remains invariant to this scalar modulation.

\section{Ablation on Distillation Targets}
\label{app:prediction_targets}

In the original DMD framework \cite{yin2024one}, the distillation objective is formulated in the clean data ($x_0$) space. To stabilize optimization across different noise levels, DMD employs a specific dynamic weighting function $w_t$:
\begin{equation}
    w_t = \frac{\sigma_t^2}{\alpha_t} \frac{CS}{\|\mu_{\text{base}}(x_t, t) - x\|_1},
\end{equation}
where $C$ is the number of channels, $S$ is the number of spatial locations, $\mu_{\text{base}}(x_t, t)$ represents the model's prediction of $x_0$, and $x$ is the generated image. The term $\sigma_t^2 / \alpha_t$ transforms the score matching objective into an $x_0$-regression loss, while the L1-norm term serves to normalize the gradient magnitude relative to the prediction error.

We investigate whether adapting this weighting strategy to our Flow Matching framework yields benefits over direct velocity regression. We train two variants of the SiT-XL/2+REPA model on ImageNet-256 for 40 epochs:
1.  \textbf{DMD-style}: Converting the velocity output $v_t$ to $x_0$ and applying the weighting $w_t$ to the loss.
2.  \textbf{FSD (Ours)}: Directly minimizing the difference in the native velocity space $(v_{\text{fake}} - v_{\text{real}})$.

Results show that the DMD-style adaptation achieved an FID of \textbf{1.52}, whereas our direct velocity regression achieved a superior FID of \textbf{1.43}. This confirms that for Rectified Flow models, supervising the native velocity field is more effective than transforming the problem into the $x_0$-space, even with adaptive weighting.

\section{Attempts at Other Discriminator Architectures}
\label{app:negative_results}

In developing the discriminator for DiT-based distillation, we explored several alternative architectures beyond our proposed MLP-on-Frozen-Teacher design. We report these attempts here for completeness; none improved performance over our baseline.

\textbf{1. Spectral Normalization.}
We attempted to replace the standard Linear layers and LayerNorm in our discriminator head with Spectral Normalization (SN), a standard technique in GANs. We applied SN to both the projection layers and the final classification head. This configuration led to slower convergence and inferior final FID compared to the standard LayerNorm formulation.

\textbf{2. Pre-trained Block Initialization.}
We experimented with initializing the discriminator head using a copy of a pre-trained DiT block (e.g., Block 14) from the teacher model, rather than a randomly initialized MLP. Fine-tuning this heavier transformer block proved unstable compared to training a lightweight MLP head from scratch.

\textbf{3. 2D Convolutional Heads.}
We investigated projecting the DiT's latent tokens back into 2D spatial grids ($H \times W$) and processing them with a stack of strided 2D Convolutions and GroupNorm layers (mimicking standard CNN discriminators). Despite the spatial inductive bias, this design yielded suboptimal results compared to the token-wise MLP approach.

\textbf{4. Discrimination on Velocity Output.}
We also attempted to apply the discriminator directly to the model's final output (the predicted velocity field $v_t$) instead of intermediate features. This approach failed to provide effective guidance, likely due to the difficulty of distinguishing real/fake patterns in the high-dimensional velocity space across different noise levels.

\section{Broader impacts}
\label{app:broader_impact}
This paper presents work aimed at advancing the efficiency of generative models. By enabling high-quality one-step generation, our approach significantly reduces the computational cost and energy consumption associated with diffusion models. While accelerating content synthesis inherits the general ethical considerations of generative AI (e.g., potential misuse for deepfakes), our work does not introduce specific new societal consequences that require separate highlighting beyond those already known in the field. 

\section{Additional Qualitative Results}
\label{app:more_samples}

\begin{figure}[h]
    \centering
    \includegraphics[width=1.0\linewidth]{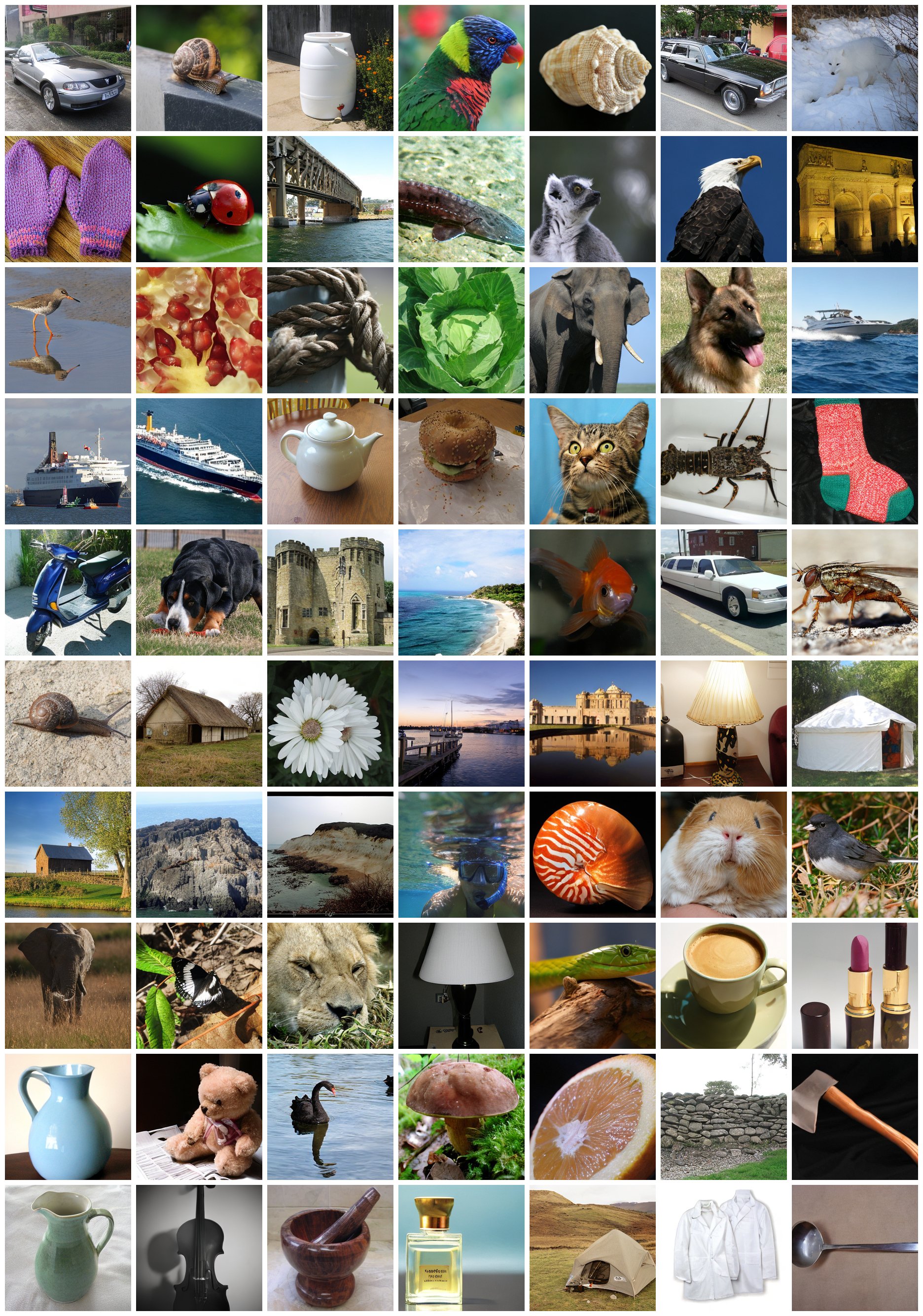} 
    \caption{\textbf{Additional One-Step Samples on ImageNet $256 \times 256$.} Selected samples generated by EASD. The results demonstrate that our method achieves photorealistic quality and covers diverse modes across various categories.}
    \label{fig:more_samples}
\end{figure}

\begin{figure}
    \centering
    \includegraphics[width=1.0\linewidth]{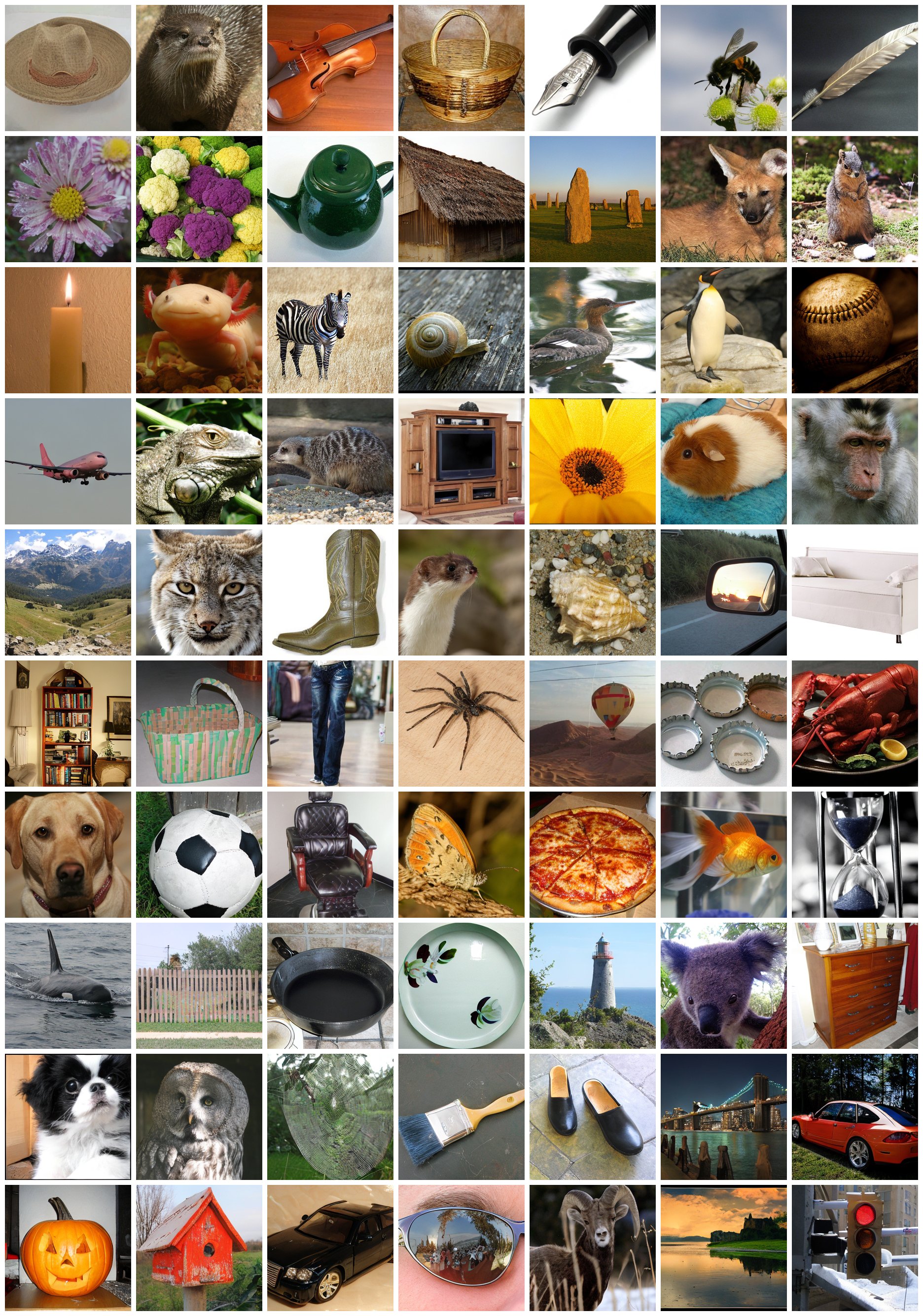}
    \caption{\textbf{Additional One-Step Samples on ImageNet $256 \times 256$.} Selected samples generated by EASD. The results demonstrate that our method achieves photorealistic quality and covers diverse modes across various categories.}
        \label{fig:more_samples2}
\end{figure}

We present additional samples generated by our EASD model trained on Class-Conditional ImageNet $256 \times 256$. 
All images are synthesized in a \textbf{single inference step (NFE=1)}. 
As shown in \cref{fig:more_samples,fig:more_samples2}, our method successfully generates high-fidelity images with correct semantics and diverse structures across a wide spectrum of classes, ranging from fine-grained textures (e.g., fur, feathers) to complex global geometries (e.g., architecture, landscapes).

\newpage
\newpage
\appendix
\onecolumn
\section*{NeurIPS Paper Checklist}

\begin{enumerate}

\item {\bf Claims}
    \item[] Question: Do the main claims made in the abstract and introduction accurately reflect the paper's contributions and scope?
    \item[] Answer: \answerYes{} 
    \item[] Justification: The abstract and introduction accurately reflect the paper’s contributions and scope.

\item {\bf Limitations}
    \item[] Question: Does the paper discuss the limitations of the work performed by the authors?
    \item[] Answer: \answerYes{} 
    \item[] Justification: The paper discuss the limitations of the work in \cref{app:limitation}.

\item {\bf Theory assumptions and proofs}
    \item[] Question: For each theoretical result, does the paper provide the full set of assumptions and a complete (and correct) proof?
    \item[] Answer: \answerYes{} 
    \item[] Justification: Detailed derivations are provided for  theoretical results: \cref{eq:gan_score_relation} (\cref{app:gan_score_derivation}), \cref{eq:dmd_grad}, and \cref{subsec:cfg_compounding}(\cref{app:recursive_guidance}).

    \item {\bf Experimental result reproducibility}
    \item[] Question: Does the paper fully disclose all the information needed to reproduce the main experimental results of the paper to the extent that it affects the main claims and/or conclusions of the paper (regardless of whether the code and data are provided or not)?
    \item[] Answer: \answerYes{} 
    \item[] Justification: All details can be found in \cref{subsec:experiment setup} and \cref{app:implementation}.

\item {\bf Open access to data and code}
    \item[] Question: Does the paper provide open access to the data and code, with sufficient instructions to faithfully reproduce the main experimental results, as described in supplemental material?
    \item[] Answer: \answerYes{} 
    \item[] Justification: All details can be found in \cref{subsec:experiment setup} and \cref{app:implementation}. Our code will also be released as open-source.

\item {\bf Experimental setting/details}
    \item[] Question: Does the paper specify all the training and test details (e.g., data splits, hyperparameters, how they were chosen, type of optimizer) necessary to understand the results?
    \item[] Answer: \answerYes{} 
    \item[] Justification: The paper specify the training and test details necessary to understand the results.

\item {\bf Experiment statistical significance}
    \item[] Question: Does the paper report error bars suitably and correctly defined or other appropriate information about the statistical significance of the experiments?
    \item[] Answer: \answerNo{} 
    \item[] Justification: Following the standard protocols in prior studies, we evaluate our model on a sufficiently large test set of 50k images. Given the observed stability of the results across such a large-scale evaluation, error bars are not provided.

\item {\bf Experiments compute resources}
    \item[] Question: For each experiment, does the paper provide sufficient information on the computer resources (type of compute workers, memory, time of execution) needed to reproduce the experiments?
    \item[] Answer: \answerYes{} 
    \item[] Justification: The experiments compute resources are provided in Appendix~\cref{app:efficiency}.
    
\item {\bf Code of ethics}
    \item[] Question: Does the research conducted in the paper conform, in every respect, with the NeurIPS Code of Ethics \url{https://neurips.cc/public/EthicsGuidelines}?
    \item[] Answer: \answerYes{} 
    \item[] Justification: Yes.

\item {\bf Broader impacts}
    \item[] Question: Does the paper discuss both potential positive societal impacts and negative societal impacts of the work performed?
    \item[] Answer: \answerNA{} 
    \item[] Justification: As stated in \cref{app:broader_impact}, our research aims to contribute to the general development of deep learning. No significant broader social or ethical implications are expected.
    
\item {\bf Safeguards}
    \item[] Question: Does the paper describe safeguards that have been put in place for responsible release of data or models that have a high risk for misuse (e.g., pre-trained language models, image generators, or scraped datasets)?
    \item[] Answer: \answerNA{} 
    \item[] Justification: The paper poses no such risks.

\item {\bf Licenses for existing assets}
    \item[] Question: Are the creators or original owners of assets (e.g., code, data, models), used in the paper, properly credited and are the license and terms of use explicitly mentioned and properly respected?
    \item[] Answer: \answerYes{} 
    \item[] Justification: Yes.

\item {\bf New assets}
    \item[] Question: Are new assets introduced in the paper well documented and is the documentation provided alongside the assets?
    \item[] Answer: \answerNA{} 
    \item[] Justification: Although the source code is not included in the initial submission, we have strictly followed all requirements and guidelines. We are committed to making the code publicly available upon acceptance/publication.

\item {\bf Crowdsourcing and research with human subjects}
    \item[] Question: For crowdsourcing experiments and research with human subjects, does the paper include the full text of instructions given to participants and screenshots, if applicable, as well as details about compensation (if any)? 
    \item[] Answer: \answerNA{} 
    \item[] Justification: The paper does not involve crowdsourcing nor research with human subjects.

\item {\bf Institutional review board (IRB) approvals or equivalent for research with human subjects}
    \item[] Question: Does the paper describe potential risks incurred by study participants, whether such risks were disclosed to the subjects, and whether Institutional Review Board (IRB) approvals (or an equivalent approval/review based on the requirements of your country or institution) were obtained?
    \item[] Answer: \answerNA{} 
    \item[] Justification: The paper does not involve crowdsourcing nor research with human subjects.

\item {\bf Declaration of LLM usage}
    \item[] Question: Does the paper describe the usage of LLMs if it is an important, original, or non-standard component of the core methods in this research? Note that if the LLM is used only for writing, editing, or formatting purposes and does \emph{not} impact the core methodology, scientific rigor, or originality of the research, declaration is not required.
    \item[] Answer: \answerNA{} 
    \item[] Justification: The core method development in this research does not involve LLMs as any important, original, or non-standard components.

\end{enumerate}

\end{document}